\documentclass[runningheads]{llncs}

\usepackage[utf8]{inputenc}
\usepackage{graphicx}
\usepackage{hyperref}
\usepackage{url}

\usepackage[numbers]{natbib}
\usepackage{amsmath}
\usepackage{amssymb}
\usepackage{mathtools}

\usepackage{algorithm}
\usepackage{algorithmic}

\usepackage{booktabs}
\usepackage{multirow}

\usepackage{eqparbox}
\usepackage[misc]{ifsym}

\begin{document}

\title{iSAGE: An Incremental Version of SAGE for Online Explanation on Data Streams
}

\titlerunning{iSAGE: An Incremental Version of SAGE}

\author{Maximilian Muschalik\inst{1}$^{,\dagger}$$^{,\text{\Letter}}$ 
\and
Fabian Fumagalli\inst{2}$^{,\dagger}$ 
\and
Barbara Hammer\inst{2} 
\and Eyke Hüllermeier\inst{1} 
}

\authorrunning{M. Muschalik et al.}

\institute{LMU Munich, MCML Munich, Geschwister-Scholl-Platz 1, Munich, Germany \and
Bielefeld University, CITEC, Inspiration 1, Bielefeld, Germany
\\$^\dagger$ denotes equal contribution
\\\Letter \ \email{maximilian.muschalik@ifi.lmu.de}}

\maketitle            

\begin{abstract}
Existing methods for explainable artificial intelligence (XAI), including popular feature importance measures such as SAGE, are mostly restricted to the batch learning scenario. However, machine learning is often applied in dynamic environments, where data arrives continuously and learning must be done in an online manner. Therefore, we propose iSAGE, a time- and memory-efficient incrementalization of SAGE, which is able to react to changes in the model as well as to drift in the data-generating process. We further provide efficient feature removal methods that break (interventional) and retain (observational) feature dependencies. Moreover, we formally analyze our explanation method to show that iSAGE adheres to similar theoretical properties as SAGE. Finally, we evaluate our approach in a thorough experimental analysis based on well-established data sets and data streams with concept drift. 
\end{abstract}

\section{Introduction}


\begin{figure}[t]
    \centering
    \includegraphics[width=\textwidth]{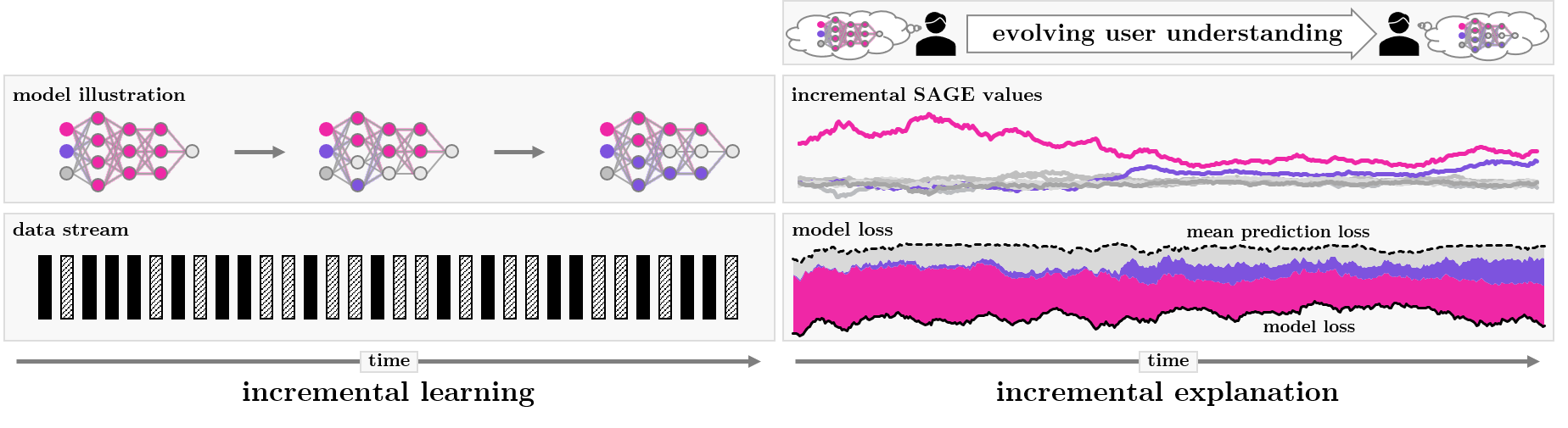}
    \caption{An incremental model is fitted on a data stream. Incrementally explaining this model with iSAGE efficiently distributes the FI scores according to the model's loss evolving the user understanding of the model over time.}
\label{fig:intorduction_image}
\end{figure}

If machine learning is used for high-stake decision-making, e.g., in healthcare \cite{Ta.2016} or energy consumption analysis \cite{GarciaMartin.2019}, models learned on data should be transparent and explainable. However, as the best performing models are often opaque in nature, this is typically not the case. The field of explainable artificial intelligence (XAI) addresses this problem by developing methods to uncover the inner working of black box models and to make the input-output relationships represented by such models more understandable \cite{Adadi.2018}.
Notably, this includes \emph{global feature importance} (global FI) methods, which quantify the influence of individual input features on the model predictions, and seek to rank the features in terms of their importance.

So far, XAI has mainly focused on static learning scenarios, where a single model is learned from data in a batch mode.  
However, in modern machine learning applications such as online credit risk scoring for financial services \cite{Jillian2020}, intrusion detection in networks \cite{Buse2018}, or sensor network analysis \cite{Bahri.2021,Davari.2021}, data is not static but coming in the form of a continuously evolving stream of data. 
In applications of that kind, online algorithms are needed for learning in an incremental mode, processing data in a sequential manner one by one. Incremental learning should not only be time- and memory-efficient, but must also account for possible changes in the underlying data distribution, which is referred to as \emph{concept drift}. Such drift may occur in different forms and for different reasons, e.g., as a change of energy consumption patterns or hospital admission criteria due to pandemic-induced lockdowns \cite{Duckworth.2021}.

In dynamic scenarios, where models are constantly evolving and reacting to their changing environment, static explanations do no longer suffice. 
Instead, explanations for monitoring dynamic models should be updated in a continuous manner, similar to the models themselves. 
In this work, we compute global FI in an incremental manner, thereby also addressing the challenge of drifting data distributions, where batch methods are likely to yield wrong explanations (cf. Fig.~\ref{fig:static_sage_in_dynamic_setting} in Appendix~\ref{sec:appendix-batch-sage}).
Providing an incremental global FI method comes with various challenges, not only conceptually and algorithmically, but also computationally, especially because the computation of many FI measures is already prohibitive in the batch setting.

\paragraph{Contribution.} We take a first step towards efficient explanations for changing models on data streams and contribute:
\begin{itemize}
    \item \emph{iSAGE}; a model-agnostic global FI algorithm that provides time- and memory-efficient incremental estimates of SAGE values and is able to react to changes in the model and concept drift. 
    \item \emph{interventional and observational iSAGE}; two conceptual approaches to define SAGE values that extend on the existing discussion of appropriate feature removal techniques 
    with an efficient incremental algorithm.
    \item \emph{open source implementation}; a well-tested and general implementation
    of our algorithms and experiments that integrates into the well-known \emph{River} \cite{Montiel.2020} Python framework.\footnote{iSAGE is implemented in \texttt{iXAI} at \url{https://github.com/mmschlk/iXAI}.}
\end{itemize}

\paragraph{Related Work.}
Global FI is an active part of XAI research, and various methods have been proposed \cite{Covert.2020}.
Model-specific methods were developed based on the magnitude of weights for linear models and neural networks \cite{Guyon_Weston_Barnhill_Vapnik_2002,Horel_Mison_Xiong_Giesecke_Mangu_2018}, as well as split heuristics for tree-based models \cite{Hastie_Tibshirani_Friedman_2009}.
Another common approach to global FI is to aggregate local explanations, such as model-agnostic LIME \cite{Ribeiro.2016} and SHAP \cite{Lundberg.2017} or neural network specific methods \cite{Sundararajan_Taly_Yan_2017,Zeiler.2014,Shrikumar_Greenside_Kundaje_2017,Springenberg_Dosovitskiy_Brox_Riedmiller_2015,Binder_Montavon_Lapuschkin_Müller_Samek_2016}.
Permutation Feature Importance (PFI) \cite{Breiman.2001} is a well-established model-agnostic, global FI method with various extensions \cite{Molnar.2020,Casalicchio.2019,konig2021relative}.
SAGE is based on the Shapley value \cite{Shapley.1953}, similar to SHAP \cite{Lundberg.2017} and LossSHAP \cite{Lundberg.2020} and overcomes computational limitations of aggregating local SHAP explanations.
Retricting a model to compute FI is done either by retaining (\emph{observational}) or breaking (\emph{interventional}) feature dependencies, where it was shown that both methods generate different explanations and the choice should depend on the application \cite{Frye.2021,Aas.2021,Chen.2020}.

Traditionally, XAI focuses on the batch learning scenario.
However, recently more methods that natively support incremental, dynamic learning environments are proposed.
For instance, online feature selection methods compute FI periodically \cite{Barddal.2019,Yuan.2018}.
Haug et al. \cite{Haug.2022} propose a concept drift detection algorithm based on clusterings and changes in SHAP's base value.
A model-specific approach for tree-based models is measuring the mean decrease in impurity (MDI) \cite{Cassidy.2014,Gomes.2019}.
In the notion of explaining change \cite{Muschalik.2022}, iPFI \cite{Fumagalli.2022} is a related model-agnostic approach that computes the traditional PFI \cite{Breiman.2001} in an incremental manner. 
To efficiently restrict the model \cite{Covert.2021}, we rely on geometric sampling \cite{Fumagalli.2022} (interventional) and a combination of the conditional subgroup approach \cite{Molnar.2020} and the TreeSHAP methodology \cite{Lundberg.2020} (observational).

Existing online FI methods are either model-specific or interpretation of the resulting feature importance scores is unintuitive, emphasizing the need for incremental variants of Shapley-based explanations, such as SAGE.

\section{Shapley Additive Global Importance (SAGE)}
\label{sec:background}
Many feature importance techniques have been proposed in recent years \cite{Covert.2021}, where each method allows to assess an importance ranking of the features.
However, interpreting the exact scores and quantifying the difference between the importance of features remains unintuitive in many cases.
Shapley-based explanations have attracted a lot of attention due to their unique mathematical properties, in particular the efficiency condition that ensures that the sum of these values over all features equals a specified model property, referred to as \emph{model behavior} \cite{Covert.2021}.
SHapley Additive Global Importance (SAGE) \cite{Covert.2020} is a well-known Shapley-based explanation technique that quantifies global FI as the contribution of individual features to the model's loss.
SAGE is further a \emph{model-agnostic} method that only relies on model evaluations and does not make any assumption about the inherent structure. 
In the following, we distinguish between the SAGE values $\phi$, a statistical concept to define Shapley-based global FI, and the SAGE estimator $\hat\phi^{\text{SAGE}}$, an efficient approximator of the SAGE values.
For a model $f: \mathcal X \to \mathcal Y$, the SAGE values $\phi(i)$ for every feature $i \in D$ are constructed, such that the sum is equal to the expected improvement in loss over using the mean prediction $\bar y := \mathbb{E}_X[f(X)]$, i.e. 
\begin{equation*}
    \nu(D) := \underbrace{\mathbb{E}_{Y}\left[\ell(\bar y,Y)\right]}_{\text{{\tiny no feature information}}} - \underbrace{\mathbb{E}_{(X,Y)}\left[\ell(f(X),Y)\right]}_{\text{{\tiny with feature information}}} = \sum_{i \in D} \phi(i),
\end{equation*}
where $\ell$ is a suitable loss (e.g, cross-entropy for classification, absolute error for regression, or kendall tau for rankings) and $(X,Y)$ refers to the joint distribution of the data-generating random variables $X$ and $Y$.
The quantity $\nu(D)$ is viewed as the improvement in loss, if \emph{all features} $D$ are known to the model.
It is then also natural to define $\nu(\emptyset)=0$, i.e. the improvement in loss is expected to be zero, if \emph{no features} are known to the model.
To quantify the importance of single features, the expected improvement in loss, if only a subset $S \subset D$ of features is known, is introduced.
To restrict this loss, the model is restricted to a subset of features $S \subset D$, by randomizing the features in $D \setminus S$.
In the following, we write $f(x) = f(x^{(S)},x^{(\bar S)})$ to distinguish the features of $x$ in $S$, $x^{(S)}$, and the features of $x$ in $\bar S := D\setminus S$, $x^{(\bar S)}$.
To randomize the features in $\bar S$, we introduce the notation $f(x,S)$ with a set $S \subset D$ and the \emph{observational} approach \cite{Lundberg.2017,Covert.2020}
\begin{equation*}
    f^{\text{obs}}(x,S) := \mathbb{E}\left[f(x^{(S)},X^{(\bar S)}) \mid X^{(S)} = x^{(S)} \right]
\end{equation*}
and the \emph{interventional} approach \cite{Chen.2020,Janzing.2020}
\begin{equation*}
   f^{\text{int}}(x,S) := \mathbb{E}\left[f(x^{(S)},X^{(\bar S)})\right].
\end{equation*}
The essential difference between the two approaches is that $f^{\text{int}}$ breaks the dependence between the features in $S$ and $\bar S$.
The observational and interventional approach are also referred to as \emph{on-manifold} and \emph{off-manifold} explanation \cite{Frye.2021}, or \emph{conditional} and \emph{marginal} expectation \cite{Janzing.2020}, respectively.
While $f^{\text{int}}$ is easy to approximate using the marginal distribution of the observed data points, approaches using $f^{\text{obs}}$ rely on further assumptions on the conditional distribution \cite{Lundberg.2017,Aas.2021}.
SAGE values are introduced using the observational approach but the SAGE algorithm relies on the interventional approach for approximation, i.e. assuming feature independence \cite{Covert.2020}. 
It was shown that both approaches yield significantly different explanations, if features are correlated \cite{Chen.2020,Frye.2021,Janzing.2020}.
We thus propose an algorithm for each approach and leave the choice of explanation to the practitioner, as it was concluded that this choice depends on the application scenario \cite{Chen.2020}.
We define the restricted improvement in loss as
\begin{equation*}
    \nu(S) := \mathbb{E}_{Y}[\ell(\bar y,Y)] - \mathbb{E}_{(X,Y)}\left[\ell(f(X,S),Y)\right] \text{ for } f \in \{f^{\text{int}},f^{\text{obs}}\}.
\end{equation*}
Then, $\nu: \mathcal P(D) \to \mathbb{R}$ defines a function over the powerset $\mathcal P(D)$, known as set function in cooperative game theory.
The SAGE values \cite{Covert.2020} are then defined as the Shapley value \cite{Shapley.1953} of $\nu$, i.e. the \emph{fair} attribution of $\nu(D)$ to individual features given its axiomatic propoerties.
\begin{definition}[SAGE values \cite{Covert.2020}]
The SAGE values are defined as
\begin{equation*}
\phi(i) := \sum_{S\subset D \setminus \{i\}} \frac 1 d \binom{d-1}{\vert S \vert}^{-1} \left[\nu(S \cup \{i\}) - \nu(S)\right].
\end{equation*}
We refer to the \emph{interventional} and \emph{observational} SAGE values, if $f^{\text{int}}$ and $f^{\text{obs}}$ are used for $f$ in $\nu$, respectively.
\end{definition}

Due to the exponential complexity of the Shapley value, the SAGE estimator uses a Monte-Carlo approximation \cite{Castro.2009} based on the representation
\begin{align*}
   \phi(i) &= \frac 1 {d!} \sum_{\pi \in \mathfrak S_D}\nu(u_i^+(\pi))-\nu(u_i^-(\pi)) = \mathbb{E}_{\pi \sim \text{unif}(\mathfrak S_D)}[\nu(u_i^+(\pi))-\nu(u_i^-(\pi))],
\end{align*}
where $\mathfrak S_D$ is the set of permutations over $D$ and $u_i^+(\pi)$ and $u_i^-(\pi)$ refer to the set of indices preceding feature $i$ in $\pi$, in- and exclusively $i$.
Plugging in the definition of $\nu$ and using Monte-Carlo estimation, the SAGE estimator is constructed.

\begin{definition}[SAGE Estimator \cite{Covert.2020}]
Given data points $(x_n,y_n)_{n=1,\dots,N}$ and permutations $(\pi)_{n=1,\dots,N}\sim \text{unif}(\mathfrak{S}_D)$ the SAGE estimator is defined as
\begin{align*}
    \hat\phi^{\text{SAGE}}(i) := \frac 1 N \sum_{n=1}^N & \ell(\hat f(x_n,u_i^-(\pi_n)),y_n) - \ell(\hat f(x_n,u_i^+(\pi_n),y_n)),
\end{align*}
with $\hat f(x,\emptyset) := \frac 1 N \sum_{n=1}^N f(x_n)$ and $\hat f(x,S) := \frac 1 M \sum_{m=1}^M f(x^{(S)},\tilde x_m^{(\bar S)})$ for $\emptyset \neq S \subset D$ with $\tilde x_m$ sampled uniformly from $x_1,\dots,x_N$.
\end{definition}

The mean prediction $\hat f(x,\emptyset)$ thereby differs to ensure that the SAGE values sum to the improvement in loss.
For each permutation $\pi_n$ and observation $(x_n,y_n)$, the SAGE estimator can be efficiently computed by iterating through the permutation and evaluating $\nu$ on the preceding elements \cite{Castro.2009,Covert.2020}.
The permutation sampling approach ensures that the efficiency condition of the Shapley value is maintained and thus the SAGE estimates sum approximately to $\nu(D)$.
In contrast to other global FI measures, where interpretation of the scores are unintuitive, SAGE yields a meaningful axiomatic interpretation.


\section{Incremental Global Feature Importance}
In the following, we consider a data stream, where at time $t$ the observations $(x_0,y_0),\dots,(x_t,y_t)$ have been observed.
On this data stream, a model $f_t$ is incrementally learned over time by updating $f_t \to f_{t+1}$ using the observation $(x_t,y_t)$. \cite{Bahri.2021,losing.2018}
Our goal is to estimate the (time-dependent) SAGE values $\phi_t$ alongside the incremental learning process using minimal resources.
In particular, in an online learning scenario, where the model is constantly adapting, huge changes in global FI scores can occur, as has been observed in Haug et al. \cite{Haug.2022} and Fumagalli et al. \cite{Fumagalli.2022}.
To guarantee the reliability of the learned models, it is crucial to understand these global FI scores over time.
The main challenge in estimating the SAGE values in an online learning scenario is that the model $f_t$ and the data-generating random variables $(X_t,Y_t)$ change over time and access to observations to compute $\hat f_t$ is limited.

While the SAGE estimator provides efficient estimates of static SAGE values for a given dataset, it does not react properly to changes in the model or concept drift.
In Appendix~\ref{sec:appendix-batch-sage}, we show an example (Fig.~\ref{fig:static_sage_in_dynamic_setting}) which illustrates that the SAGE estimator yields wrong importance scores if the underlying distribution or model is not static. 
Furthermore, computing the SAGE estimator repeatedly in an incremental setting on a data stream quickly becomes infeasible.
As a remedy, we propose incremental SAGE (iSAGE), an incremental estimator, which reacts to changing distributions and is able to explain dynamic, time-dependant models.
To compare iSAGE in an incremental learning setting, we first propose Sliding Window SAGE (SW-SAGE), a time-sensitive baseline estimator that repeatedly computes the SAGE estimator on a sliding window.

\subsubsection{Sliding Window SAGE (SW-SAGE)}
A naive approach of approximating SAGE values in an incremental manner is through repeated calculation within a sliding window (SW), which we denote as \emph{SW-SAGE}.
Applying SW-SAGE, necessitates storing all historical observations $(x_{t},y_{t})$ for the last $w$ (window length) observations, and recomputing the SAGE estimator from scratch based on the most up-to-date model $f_{t}$.
The main computational effort of SW-SAGE stems from evaluating the model $f_t$ and thus scales linearly with the window length $w$.
The size $w$ of the window has a profound effect on the resulting SAGE estimates.
Choosing a large value for $w$, may increase the quality of the estimated SAGE values (larger sample), but can also lead to wrong importance scores, since the window may contain outdated observations.
Vice versa, a window size too small leads to a high variance.
\hfill

\begin{algorithm}[t]
{\footnotesize
\caption{Incremental SAGE (iSAGE)}
\label{alg:incremental-sage}
\begin{algorithmic}[1]
\REQUIRE stream $\left\{ x_{t}, y_{t}\right\}_{t=1}^{\infty}$, feature indices $D = \{1, \dots, d \}$, model $f_{t}$, loss function $\ell$, and inner samples $m$ \\[0.25em] 
\STATE Initialize $\hat\phi^{1} \gets 0,\hat\phi^{2} \gets 0,\dots,\hat\phi^{d} \gets 0$, and smoothed mean prediction $y_{\emptyset} \gets 0$
\FORALL{$(x_{t},y_{t}) \in$ stream}
    \STATE Sample $\pi$, a permutation of $D$
    \STATE $S \gets \emptyset$
    \STATE $y_\emptyset \gets (1 - \alpha) \cdot y_\emptyset + \alpha \cdot f(x_{t})$ \COMMENT{Udpate mean prediction}
    \STATE $ \text{lossPrev} \gets \ell(y_\emptyset,y_{t})$ \COMMENT{Compute mean prediction loss}
    \FOR[Iterate over $\pi$]{$j = 1$ to $d$}
        \STATE $S \gets S \cup \left\{ \pi[j] \right\}$
        \STATE $y \gets 0$
        \FOR[Marginalize prediction with $S$]{$k = 1$ to $m$}
            \STATE Sample $x^{(\bar S)}_{k} \sim \mathbb{Q}^{(x,S)}_t$ \COMMENT{interventional (Algorithm~\ref{alg:marginal-feature-removal}) or observational (Algorithm~\ref{alg:incremental-tree-storage})}
            \STATE $y \gets y + f_t(x^{(S)}_{t}, x^{(\bar S)}_{k})$
        \ENDFOR
        \STATE $\bar y \gets \frac{y}{m}$
        \STATE $\text{loss} \gets \ell(\bar y,y_{t})$
        \STATE $\Delta \gets \text{lossPrev} - \text{loss}$
        \STATE $\hat \phi^{\pi[j]} \gets (1 - \alpha) \cdot \phi^{\pi[j]} + \alpha \cdot \Delta$
        \STATE $\text{lossPrev} \gets \text{loss}$
    \ENDFOR
\ENDFOR
\STATE \textbf{return} $\phi^{1},\phi^{2},\dots,\phi^{d}$
\end{algorithmic}
}
\end{algorithm}

\subsection{Incremental SAGE (iSAGE)}
The high computational effort and the inability to reuse past results, because of the dynamic nature of $f_{t}$, strictly limits SW-SAGE in many scenarios, further discussed in Section~\ref{sec:experiments-incremental}.
As a result, we now propose a time- and memory-efficient variant of SW-SAGE, which we refer to as \emph{incremental SAGE} (iSAGE).
The iSAGE algorithm computes the (time-dependent) SAGE values $\phi_t$ at time $t$ and is able to react to changes in the model and concept drift, while updating its estimates efficiently in an incremental fashion with minimal computational effort.
At each time step, we observe a sample $(x_t,y_t)$ from the data stream, and our goal is to update the estimate using the current model $f_t$.
We sample $\pi_t\sim \text{unif}(\mathfrak S_D)$ to compute the marginal contribution for $i \in D$ as
\begin{align*}
   \Delta_t(i) :=& \ell(\hat f_t(x_t,u_i^-(\pi_t)),y_t)- 
   \ell(\hat f_t(x_t,{u_i^+(\pi_t)}),y_t),
\end{align*}
where $\hat f_t(x,S)$ is a time-sensitive approximation of the restricted model, further discussed in Section \ref{sec:feature_removal_algorithms}.
These computations are then averaged over time, which yields the iSAGE estimator, outlined in Algorithm \ref{alg:incremental-sage}.
\begin{definition}[iSAGE]
The iSAGE estimator is recursively defined as
\begin{align*}
     \text{iSAGE: }\hat \phi_t(i) &=  (1-\alpha)\cdot \hat \phi_{t-1}(i) + \alpha \cdot \Delta_t(i),
\end{align*}
where $\alpha >0$ and computation starts at $0<t_0<t$ with  $\hat \phi_{t_0}(i) := \Delta_{t_0}(i)$.
\end{definition}

The iSAGE estimator thus approximates $\phi_t$ by exponentially smoothing previous SAGE estimates, as $\mathbb{E}[\Delta_t(i)]=\phi_t(i)$.
In the static batch setting, the SAGE estimator computes the restricted model $f_t(x,S)$ by sampling uniformly from observations in the dataset.
However, when $f_t$ is incrementally updated in the data stream setting, access to previous observations is limited as observations are discarded after the incremental update of the model.
Furthermore, the distribution of previous observations might change over time, so recently observed samples should be preferred.
We thus present two sampling strategies to implement the observational and interventional approach in an incremental fashion.

\subsection{Incremental Feature Removal Strategies}
\label{sec:feature_removal_algorithms}

As mentioned in Section~\ref{sec:background}, SAGE is defined using the observational approach, which is then approximated by the interventional approach, i.e. sampling from the marginal distribution and assuming feature independence.
Clearly, this constitutes a strong assumption that is rarely satisfied in practice.
Instead, we sample from the marginal distribution to compute \emph{interventional} iSAGE and propose a novel approach to compute \emph{observational} iSAGE, by approximating the conditional distribution.
This aligns with \cite{Chen.2020}, where it is claimed that the choices of feature removal is dependent on the application scenario.
For both approaches we now provide a time- and memory-efficient incremental sampling approach by maintaining time-dependent reservoirs to estimate $f(x,S)$.

\begin{definition}[Estimator for $f(x,S)$]
At time $t$, we define for $\emptyset \neq S \subset D$
\begin{equation*}
    \hat f_t(x,S) := \frac 1 M \sum_{m=1}^M f_t(x^{(S)},\tilde x^{(\bar S)}_m) \text{ with } x_1,\dots,x_M \sim \mathbb{Q}_t^{(x,S)},
\end{equation*}
where $\bar S := D \setminus S$ and $\mathbb{Q}_t^{(x,S)}$ is a sampling distribution over features in $\bar S$. Further, $\hat f_t(x,\emptyset) := (1-\alpha)f_{t-1}(x,\emptyset) + \alpha f_t(x_t)$ and $\hat f_{t_0}(x,\emptyset):=f_{t_0}(x_{t_0})$.
\end{definition}

The interventional approach breaks the feature dependency and thus $\mathbb{Q}^{(x,S)}_t$ does not depend on the location $x$, whereas for the observational $\mathbb{Q}^{(x,S)}_t$ does depend on both, the location $x$ as well as the subset $S$.
We now describe incremental sampling algorithms to sample from $\mathbb{Q}^{(x,S)}$ for either approach.

\subsubsection{Interventional iSAGE}
The interventional approach in the incremental learning setting is defined as $f_t^{\text{int}}(x,S) := \mathbb{E}\left[f_t(x^{(S)},X_t^{(\bar S)})\right]$. 
The batch SAGE algorithm samples uniformly from all observations from the given dataset.
In an incremental learning scenario, this approach has significant drawbacks.
First, access to previous observations is limited, as storing observations may be infeasible for the whole data stream.
Second, the distribution of $X_t$ may change over time, and it is, thus, beneficial to favor \emph{recent} observations over older data points.
The geometric sampling strategy, proposed by Fumagalli et al. \cite{Fumagalli.2022}, accounts for both of these challenges.
Geometric sampling maintains \emph{one} reservoir of length $L$, that is updated at each time step with an incoming data point by uniformly replacing a data point from the reservoir.
Then, at each time step, observations $\tilde x_m$ are uniformly chosen from the reservoir.
The geometric sampling strategy (fully initialized at time step $L := t_0$) thus chooses a previous observation from time $r$ at time $s$ with probability $L^{-1}(1-L^{-1})^{s-r-1}$ for $r\geq L$, which clearly favors more recent observations.
The complete procedure is given in Algorithm~\ref{alg:marginal-feature-removal}.
At any time $t$, geometric reservoir sampling requires a storage space of $\mathcal{O}(L)$ data points. 
It has been shown that the geometric sampling procedure is favorable in scenarios with concept drift compared to memory-efficient uniform sampling approaches, such as general reservoir sampling \cite{Fumagalli.2022}.

\subsubsection{Observational iSAGE}
The interventional approach can generate unrealistic observations when features are highly correlated, resulting in out-of-distribution evaluations of the model.
When understanding causal relationships, it might be inappropriate to evaluate the model outside the data manifold \cite{Chen.2020}, and we thus propose an alternative approach that can incorporate feature dependence in the incremental sampling process. The observational approach in the incremental setting is defined as $f_t^{\text{obs}}(x,S) := \mathbb{E}\left[f_t(x^{(S)},X_t^{(\bar S)}) \mid X_t^{(S)} = x^{(S)} \right]$.
While observing data points $x_t$, we train for every feature $i \in D$ an incremental decision tree that aims at predicting $x_t^{(i)}$ given the remaining feature values $x_t^{(D \setminus \{i\})}$.
We then traverse the incremental decision tree using the input $x_t$ and maintain a reservoir of length $L$ at each leaf node, using the geometric sampling strategy described above, i.e. uniformly replacing an observation in the leaf's reservoir.
This yields a reservoir of length $L$ at every leaf node of the incremental decision tree, where both, the decision tree as well as the reservoir change over time.
We propose to use a Hoeffding Adaptive Tree (HAT), a popular incremental decision tree \cite{Hulten.2001}, to adaptively maintain the structure.
The approach can be viewed as an incremental variant of the \emph{conditional subgroup} approach \cite{Molnar.2020}.
Given a subset $S \subset D$ and an observation $x_t$, we obtain the values of $\tilde x_m^{(\bar S)}$ separately for each feature $j \in \bar S$.
Using $x_t$, we traverse the HAT and at every decision node that splits on a feature in $\bar S$, we randomly split according to the split ratio of previous observed inputs, a statistic that is inherently available for a HAT.
From the reservoir at the resulting leaf node, we then uniformly sample values for $\tilde x_m^{(j)}$ and repeat this process for every feature $j \in \bar S$ until we obtain all values for $\tilde x_m^{(\bar S )}$.
This methodology parallels the TreeSHAP approach of traversing decision trees for absent features, referred to as path dependent TreeSHAP \cite{Lundberg.2020}.
Notably, our approach allows to extend the conditional subgroup approach to an arbitrary feature subset $S \subset D$ while maintaining only \emph{one} decision tree \emph{per feature} and further extends the approach to an incremental setting.
The observational approach via HAT has a space complexity of $\mathcal{O}(d \cdot T^R\cdot L)$ where $R$ refers to the HATs' maximum tree depth, $T$ is the maximum number of tree splits, and $L$ is the size of the reservoir at each leaf node.

\subsection{Approximation Guarantees for Static Environments}
We presented iSAGE as a time- and memory-efficient algorithm to estimate SAGE values over time incrementally.
In contrast to the SAGE estimator, iSAGE reacts to changes in the model as well as concept drift, which we demonstrate empirically in Section \ref{sec:experiments}.
Analyzing iSAGE theoretically in an incremental learning scenario would require strong assumptions on the data-generating random variables $(X_t,Y_t)$ and the approximation quality of the learned model $f_t$, as the iid assumption in general is not fulfilled.
Instead, we now show theoretically that iSAGE has similar properties as the SAGE estimator in a static learning environment.
In the following, we assume that $f \equiv f_t$ is a constant model and $(X,Y) \equiv (X_t,Y_t)$ a stationary data generating process. 
We further assume that $\mathbb{Q}^{(x,S)}_t$ is the true marginal (interventional) or conditional (observational) distribution and that samples are drawn iid, similar to Covert et al. \cite{Covert.2020}.
\begin{theorem}\label{thm::consistency}
For iSAGE $\hat\phi_t(i) \to \phi_t(i)$ for $M\to \infty$ and $t \to \infty$.
\end{theorem}

Theorem~\ref{thm::consistency} shows that iSAGE converges to the SAGE values.
Further, the variance is controlled by $\alpha$.

\begin{theorem}\label{thm::approx-quality}
The variance of iSAGE is controlled by $\alpha$, i.e. $\mathbb{V}[\hat\phi_t(i)] = \mathcal O (\alpha)$.
\end{theorem}

Lastly, we show that iSAGE does not differ much from the SAGE estimator.

\begin{theorem}\label{thm::approx-SAGE}
Given the SAGE estimator $\hat\phi^{\text{SAGE}}_t(i)$ computed at time $t$ over all previously observed data points, it holds for iSAGE with $M\to \infty$, $\alpha = \frac 1 t$ and every $\epsilon > (1-\alpha)^{t-t_0+1}$ that $\mathbb{P}\left( \vert \hat\phi_t(i) - \hat\phi_t^{\text{SAGE}}(i) \vert > \epsilon \right) = \mathcal O(\frac 1 t)$.
\end{theorem}

While iSAGE admits similar properties as the SAGE estimator in a static environment, we showcase in our experiments that iSAGE is able to efficiently react to model changes and concept drift in an incremental learning setting.

\section{Experiments}
\label{sec:experiments}
We now utilize iSAGE in multiple experimental settings.
In Section \ref{sec:experiments-incremental}, we show how iSAGE can be efficiently applied in dynamic environments with concept drift.
In Section \ref{sec:experiments-gt}, we construct a synthetic ground-truth scneario for a data stream with concept drift and show that iSAGE is able to efficiently recover the SAGE values.
In Section \ref{sec:experiments-feature-removal}, we illustrate the difference of interventional and observational iSAGE, which yield profoundly different explanations.
In Section \ref{sec:experiments-batch}, we show that iSAGE leads to the same results as the SAGE estimator in a static environment validating our theoretical results.
As our iSAGE explanation technique is inherently model-agnostic, we train and evaluate our method on different incremental and batch models. \footnote{All model implementations are based on \textit{scikit-learn} \cite{scikit-learn}, \textit{River} \cite{Montiel.2020}, and \textit{torch} \cite{torch.2017}. 
The data sets and streams are retrieved from \textit{OpenML} \cite{OpenML2020} and \textit{River}.
The data sets are described in detail in Appendix~\ref{sec:appendix-data-set-description}. 
\href{https://github.com/mmschlk/iSAGE-An-Incremental-Version-of-SAGE-for-Online-Explanation-on-Data-Streams}{\texttt{https://github.com/mmschlk/iSAGE-} \texttt{An-Incremental-Version-of-SAGE-for-Online-Explanation-on-Data-Streams}}.}

\subsection{iSAGE in Dynamic Environments with Concept Drift}
\label{sec:experiments-incremental}

\begin{figure}[t]
    \centering
    \includegraphics[width=0.495\textwidth]{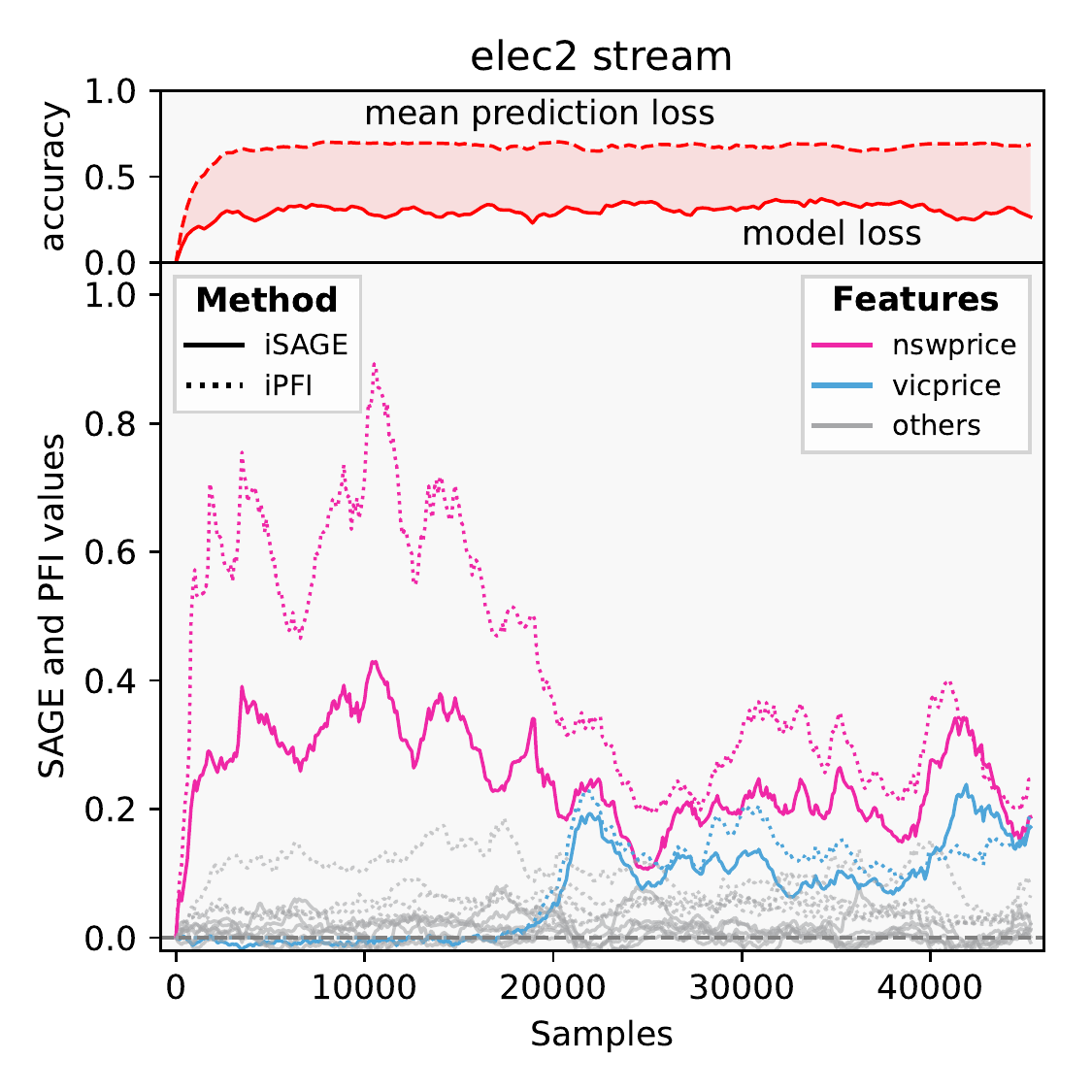}
    \includegraphics[width=0.49\textwidth]{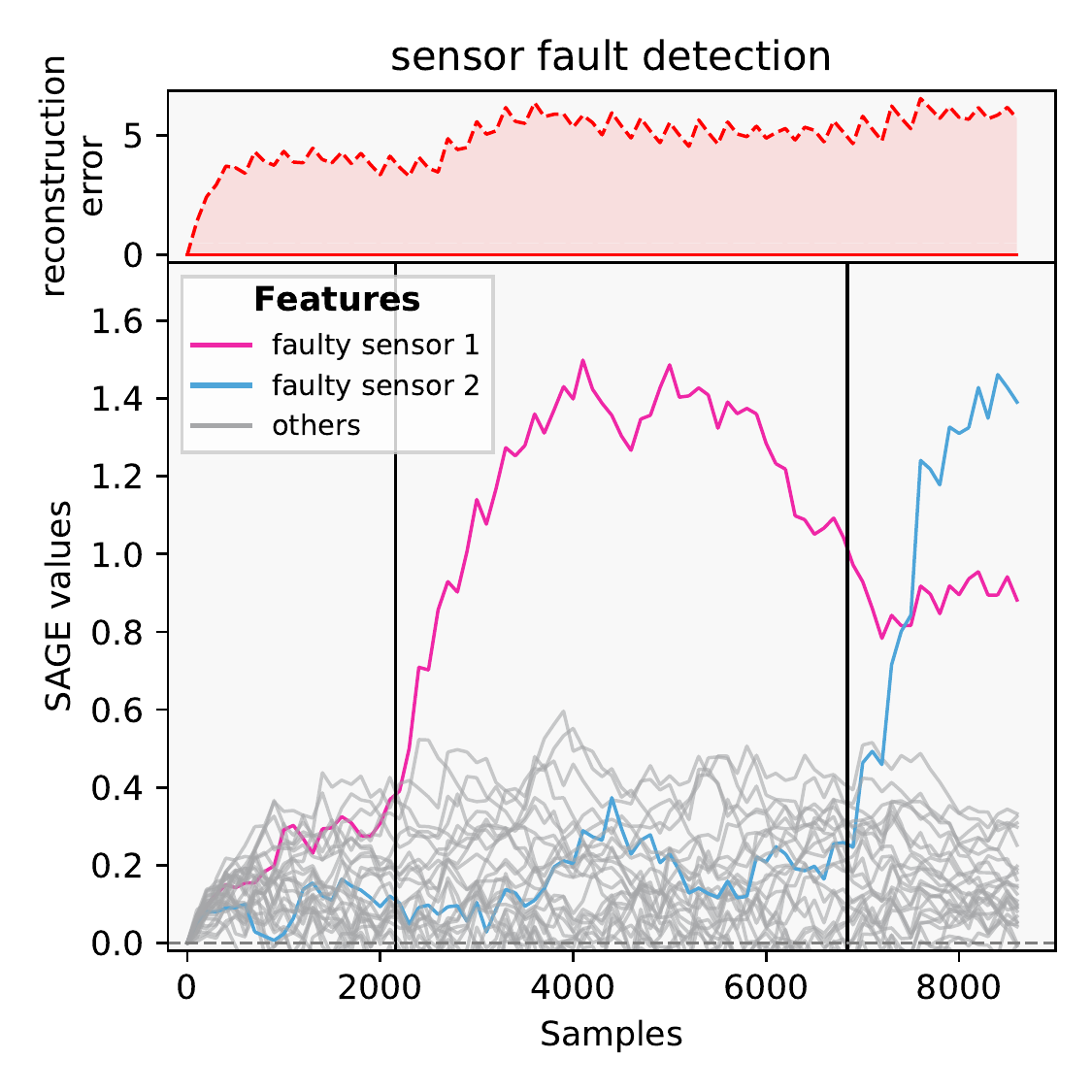}
    \caption{iSAGE and iPFI of an ARF on \emph{elec2} (left) and iSAGE for an incrementally fitted autoencoder for \emph{fault detection} based on the reconstruction loss (right).}
    \label{fig:elec2-incremental}
    \label{fig:sensor_faults_ae}
    \vspace{-1em}
\end{figure}

In this experiment, we demonstrate the explanatory capabilities of iSAGE in a dynamic learning scenario with concept drift.
We illustrate how iSAGE uncovers hidden changes in black box incremental models applied in real-world incremental learning scenarios where models are updated with every new observation.
We compare iSAGE with incremental permutation feature importance (iPFI) \cite{Fumagalli.2022}, which is up to our knowledge, the only model-agnostic explanation method that can be applied in an incremental learning setting.
For additional experiments and a comparison with the mean decrease in impurity (MDI) for tree-based models \cite{Gomes.2019}, we refer to the supplement material (cf., \ref{sec:appendix-incremental-models}).
Fig.~\ref{fig:elec2-incremental} explains the incremental learning procedure for an ARF classifier on the \emph{elec2} data stream. 
\\
Both methods detect similar feature importance rankings with varying absolute values.
In contrast to iPFI, iSAGE explanations sum to the time-dependent difference in model loss over the loss using the mean prediction, due to the efficiency axiom of the Shapley value, which naturally increases interpretability of the method.
Both methods correctly reveal the hidden changes in the model induced by the concept drift in the well-studied \emph{elec2} \cite{elec2.1999}.
The concept drift, which stems from the \emph{vicprice} feature not having any values in the first $\approx20k$ observations, would be obfuscated by solely plotting the model performance without any online explanations.

\noindent \emph{Localization of sensor faults.}
As an illustrative example, we conduct an experiment to show how online SAGE values can detect sensor faults in online sensor networks, which constitutes a challenging predictive maintenance problem \cite{Davari.2021,Vaquet.2022}.
Similar to Hinder et al. \cite{Hinder.2023}, we simulate sensor network data of water pressures including sensor faults (vertical lines in Fig.~\ref{fig:sensor_faults_ae} denote the time points) via the L-Town \cite{Vrachimis.2022} simulation tool \cite{Klise.2017} and explain online learning models. 
We incrementally fit \footnote{
We fit the autoencoder with each new data point by conducting a single gradient update (i.e., batch size of 1). For further information about the experimental setup, we refer to Appendix~\ref{sec:experiments_water_data}.
} and explain a NN autoencoder on the sensor readings.
Fig.~\ref{fig:sensor_faults_ae} shows how the autoencoder's reconstruction error is distributed onto the individual senor values by iSAGE.
Notably, the faulty sensor can easily be identified through inspection of the iSAGE values after the sensor faulted. 

\subsection{Approximation quality with synthetic ground-truths.}
\label{sec:experiments-gt}

\begin{figure*}[t]
    \centering
    \includegraphics[width=0.99\textwidth]{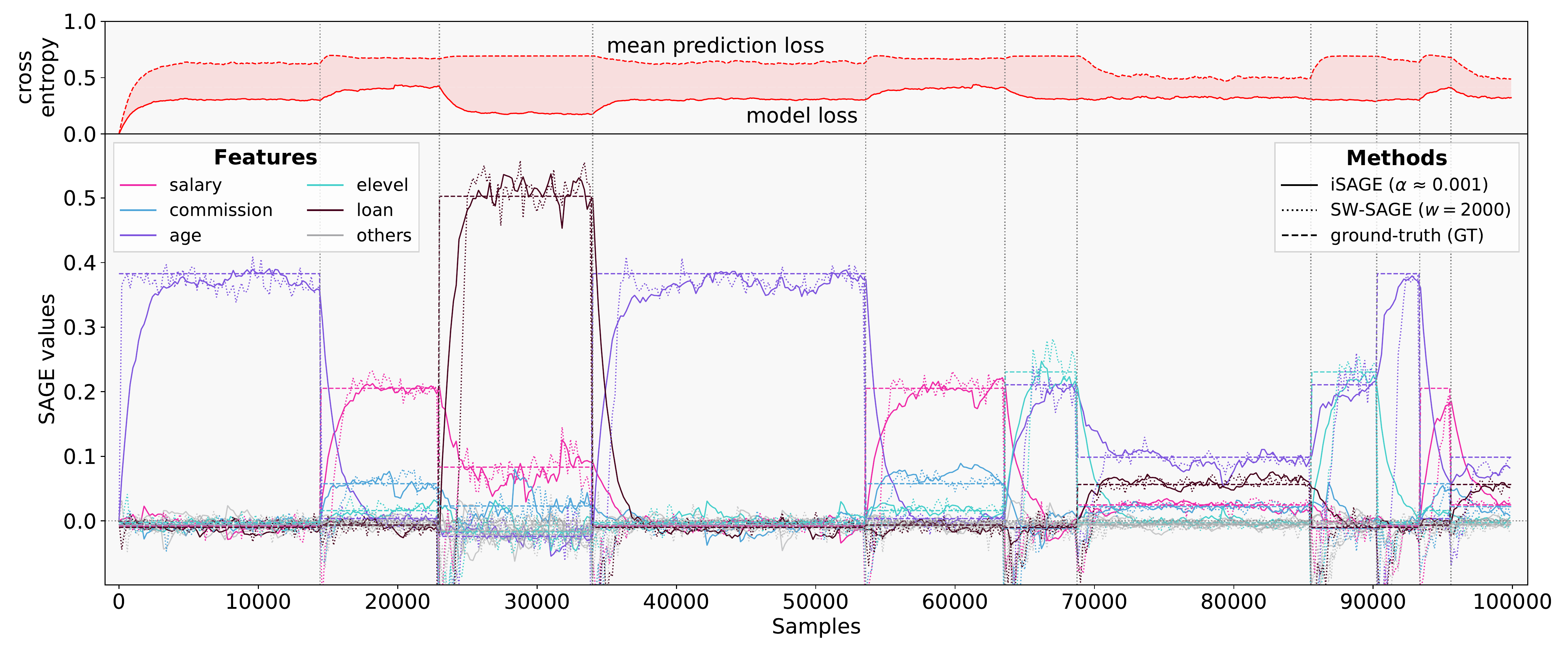}
    \caption{iSAGE (solid), SW-SAGE (dotted) and GT (dashed) values for an example GT stream. SW-SAGE is computed with a stride of $100$ ($0.05 \cdot w$) resulting in an overhead 20 times higher than iSAGE.}
    \label{fig:gt-illustration}
\end{figure*}

\begin{table}[b]
\centering
\caption{Approximation quality of iSAGE ($\text{inc}_{c}$) and SW-SAGE ($\text{SW}_{c}$) on synthetic GT data streams for $20$ iterations ($c$ denotes the factor of additional model evaluations compared to iSAGE). The complete results are given in Table~\protect\ref{tab:ground-truth-complete-results}.}
\vspace{0.25em}
\label{tab:my-table}
{
\setlength{\tabcolsep}{2pt}
\footnotesize
\begin{tabular}{@{}llcccccc@{}}
\toprule
\multicolumn{2}{l}{\textbf{scenario}} & \multicolumn{2}{c}{\textbf{high}} & \multicolumn{2}{c}{\textbf{middle}} & \multicolumn{2}{c}{\textbf{low}} \\
\multicolumn{2}{l}{\textbf{size ($w$)}} & $500$ & $1\,000$ & $500$ & $1\,000$ & $500$ & $1\,000$ \\ \midrule
\multirow{3}{*}{\begin{tabular}[c]{@{}c@{}}\textbf{MSE}\\($\sigma$)\end{tabular}} & $\text{inc}_1$ & \textbf{\begin{tabular}[c]{@{}c@{}}0.034\\(0.021)\end{tabular}} & \textbf{\begin{tabular}[c]{@{}c@{}}0.038\\(0.022)\end{tabular}} & \textbf{\begin{tabular}[c]{@{}c@{}}0.027\\(0.023)\end{tabular}} & \textbf{\begin{tabular}[c]{@{}c@{}}0.027\\(0.026)\end{tabular}} & \textbf{\begin{tabular}[c]{@{}c@{}}0.015\\(0.012)\end{tabular}} & \textbf{\begin{tabular}[c]{@{}c@{}}0.013\\(0.009)\end{tabular}} \\
 & $\text{SW}_{20}$ & \begin{tabular}[c]{@{}c@{}}0.283\\(0.262)\end{tabular} & \begin{tabular}[c]{@{}c@{}}0.420\\(0.360)\end{tabular} & \begin{tabular}[c]{@{}c@{}}0.191\\(0.271)\end{tabular} & \begin{tabular}[c]{@{}c@{}}0.320\\(0.487)\end{tabular} & \begin{tabular}[c]{@{}c@{}}0.049\\(0.043)\end{tabular} & \begin{tabular}[c]{@{}c@{}}0.078\\(0.081)\end{tabular} \\
 & $\text{SW}_1$ & \begin{tabular}[c]{@{}c@{}}0.248\\(0.198)\end{tabular} & \begin{tabular}[c]{@{}c@{}}0.462\\(0.413)\end{tabular} & \begin{tabular}[c]{@{}c@{}}0.183\\(0.200)\end{tabular} & \begin{tabular}[c]{@{}c@{}}0.399\\(0.792)\end{tabular} & \begin{tabular}[c]{@{}c@{}}0.061\\(0.067)\end{tabular} & \begin{tabular}[c]{@{}c@{}}0.080\\(0.079)\end{tabular} \\ 
\bottomrule
\end{tabular}
}
\end{table}

We compare iSAGE to the inefficient baseline SW-SAGE, as well as synthetic ground-truth (GT) values estimated using the SAGE estimator.
Conducting GT experiments in an incremental learning setting where models change with every new observation is computationally prohibitive.
Moreover, it is not defined what constitutes a GT online explanation for real-world data streams with hidden drifts.
We construct a data stream that consists of multiple sub-streams, each with different classification functions, i.e. inducing sudden concept drift when sub-streams are switched.
Within each substream, we maintain a \emph{static} pre-trained model with a pre-computed (constant) GT explanation.
We observe how differently parameterized SW-SAGE and iSAGE estimators approximate the pre-computed GT values, see Fig.~\ref{fig:gt-illustration}, and measure the approximation quality in terms of MSE and MAE.
We repeat the complete experimental setup 20 times for each frequency scenario and summarize the resulting approximation errors (MSE) in Table~\ref{tab:ground-truth-complete-results} and Fig.~\ref{fig:gt-illustration}.
Independently of the substantially increased computational overhead (up to $20$ times), SW-SAGE's approximation quality is substantially worse compared to iSAGE.
In some scenarios, SW-SAGE reaches the GT values faster than iSAGE. 
Yet, in the important phases of change, SW-SAGE' estimates are substantially worse than iSAGE's (see Fig.~\ref{fig:appendix-gt-streams-detail-view} for a detailed view).
This is a result from SW-SAGE attributing equal weight to outdated observations after a concept drift and the current model $f_t$ classifying the samples differently than the model before.
iSAGE, however, smoothly changes between the different concepts.

\subsection{Interventional and Observational iSAGE}
\label{sec:experiments-feature-removal}

In the presence of dependent variables, the choice of an interventional or observational approach has a profound effect on the SAGE values.
In this experiment, we compare both approaches using the efficient incremental algorithms presented in Section~\ref{sec:feature_removal_algorithms}.
An ARF model is trained and explained on the synthetic \emph{agrawal} data stream.
The synthetic classification function is defined in Appendix~\ref{sec:appendix-experiments-feature-removal}.
In this stream the $X_{\text{commission}}$ feature ($X_{\text{com.}}$) directly depends on $X_{\text{salary}}$.
Whenever the \emph{salary} of an applicant exceeds $75k$, no \emph{commission} is given ($X_{\text{com.}} = 0$), and otherwise the commission is uniformly distributed ($X_{\text{com.}} \sim U(10k, 75k)$).
Fig.~\ref{fig:agrawal_conditional_vs_marginal} showcases how interventional and observational iSAGE differ.
\\
No significant importance is distributed to the $X_{\text{com.}}$ feature, if observational iSAGE is used, as the information present in $X_{\text{com.}}$ can be fully recovered by the observational approach based on $X_{\text{salary}}$.
The importance is distributed onto the remaining two important features $X_{\text{salary}}$ and $X_{\text{age}}$.
However, when interventional iSAGE is used, the importance is also distributed to $X_{\text{com.}}$, as the model is evaluated outside the data manifold.
The unrealistic feature values uncover that the incremental model has picked up on the transient relationship between the target values and the feature $X_{\text{com.}}$.

\subsection{iSAGE and SAGE in Static Environments}
\label{sec:experiments-batch}

\begin{figure}[t]
\begin{minipage}[t]{0.48\textwidth}
    \centering
    \includegraphics[width=\textwidth]{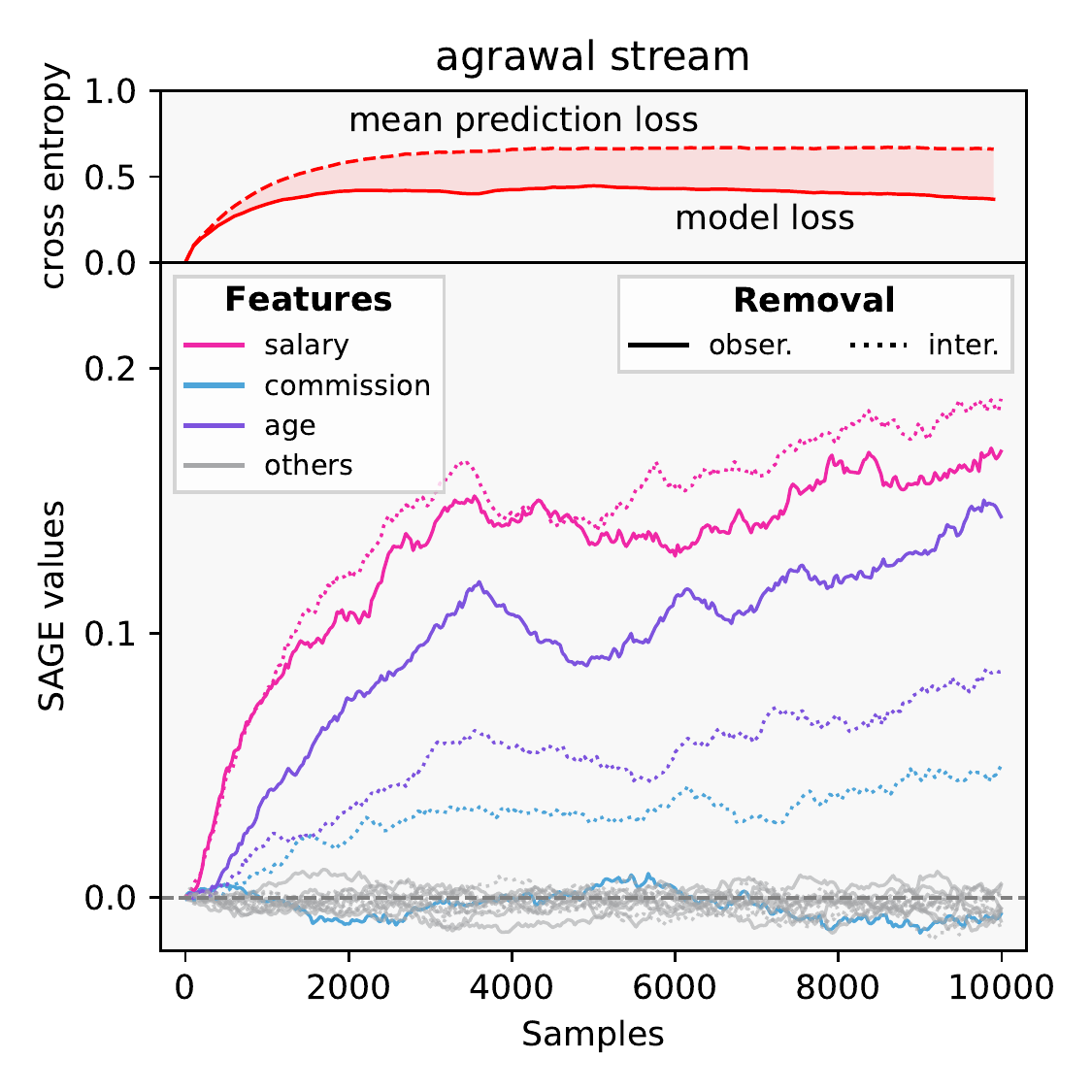}
    \caption{Interventional and observational iSAGE for an ARF on an \emph{agrawal} stream. The features have profoundly different scores.}
    \label{fig:agrawal_conditional_vs_marginal}
\end{minipage}
\hfill
\begin{minipage}[t]{0.48\textwidth}
    \centering
    \includegraphics[width=\textwidth]{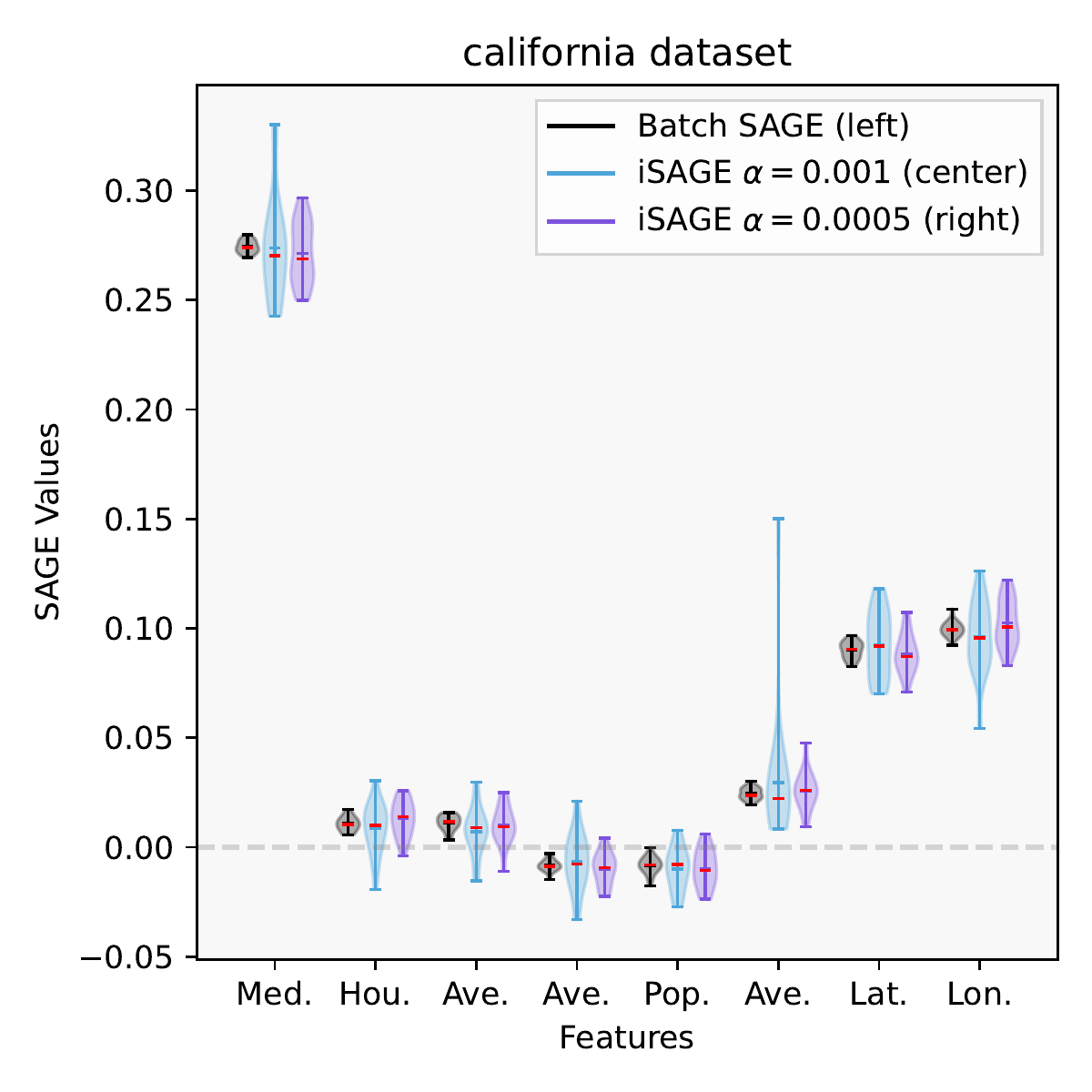}
    \caption{SAGE values (median in red) per feature of the \emph{california} dataset for the SAGE estimator (left), and interventional iSAGE (middle and right).}
    \label{fig:static_model}
\end{minipage}
\end{figure}

We consider a static learning scenario, in which we compare interventional iSAGE with Covert et al. \cite{Covert.2020}'s original SAGE approach for well-established benchmark batch datasets.
The models are pre-trained and then explained. 
We apply Gradient Boosting Trees (GBTs) \cite{Friedman.2001}, LightGBM models (LGBM) \cite{ke2017lightgbm}, and neural networks (NNs).
The original SAGE explanations are directly computed from the batch datasets.
iSAGE experiences the datasets as a randomly shuffled data stream where the model is not updated incrementally.
We run this explanation procedure 20 times and illustrate the SAGE values on the \emph{california} example dataset in Fig.~\ref{fig:static_model} (more datasets in Section~\ref{sec:appendix-results-ground-truth}).
Fig.~\ref{fig:static_model} shows that iSAGE approximates SAGE in the static setting on average with a higher variance.
The higher variance is a direct result of the iSAGE having no access to future data points and the exponential smoothing mechanism controlled by $\alpha$.
iSAGE, thus, focuses more on recent samples, which is essential for non-stationary environments like incremental learning under concept drift.

\section{Conclusion and Future Work}
We propose and analyze iSAGE, a novel and model-agnostic explanation procedure to compute global FI in dynamic environments based on time-dependent SAGE values.
In contrast to the batch SAGE algorithm \cite{Covert.2020}, iSAGE is able to efficiently react to concept drift and changes in the model.
We further extend SAGE with the observational and interventional SAGE values as distinctive objectives and present efficient incremental iSAGE variants, that are able to estimate these values over time and react to changes in the model and concept drift.
In particular, we present an incremental approximation for the observational approach that combines the conditional subgroup approach \cite{Molnar.2020} and the TreeSHAP methodology \cite{Lundberg.2020}, which could also be used in a static learning environment to further improve the SAGE algorithm.
We empirically confirm profound differences in both explanations depending on the choice of approach, which yields supporting arguments in the interventional and observational debate \cite{Frye.2021,Janzing.2020,Chen.2020} that the choice should depend on the application scenario \cite{Chen.2020}.
In a static environment, we prove that iSAGE has similar properties as SAGE and that both do not differ significantly.
We further illustrate the efficacy of incremental explanations in multiple experiments on benchmark data sets and streams and conduct a ground-truth comparison.
\\
Still, approximating Shapley values remains a computationally challenging problem.
Moreover, this approach does not address the problem of incrementally decomposing the interactions between features, which requires further investigation.
Finally, the interaction between human users and incrementally created explanations derived from methods like iSAGE need to be vigorously evaluated to identify further research opportunities. 

\section*{Acknowledgements}
We gratefully acknowledge funding by the Deutsche Forschungsgemeinschaft (DFG, German Research Foundation): TRR 318/1 2021 – 438445824.
The authors want to thank Rohit~Jagtani for supporting the implementation and valuable discussions.
The authors want to thank Gunnar~König for valuable discussions and feedback.

\bibliography{references}

\clearpage

\section{Ethical Statement}
We propose iSAGE as a novel XAI method that \emph{enables} explanations for any incrementally trained and dynamic black-box model.
This is a novel research direction, which could lead to various use cases. 
Models, that could not be evaluated before, because of computational restrictions can be investigated with iSAGE.
This enables high-performing models to be applied in various critical application domains such as healthcare \cite{Ta.2016}, energy consumption analysis \cite{GarciaMartin.2019}, credit risk scoring \cite{Jillian2020}.
These application domains could greatly benefit from XAI methods such as iSAGE, since they can help in uncovering inherent biases or problems with fairness.
This could help with more targeted regulation and scrutinization of opaque, yet high-performing, technologies than without explanations.
On the other hand, improved interpretability may also lead to an increased acceptance and exploitation of potentially harmful applications using black box models.


\appendix
\clearpage

\section*{Organization of the Supplement Material}
The proofs of our Theorems are given in Section~\ref{sec:appendix-proofs}.
Section~\ref{sec:appendix-marginal-tree-feature-removal} contains technical details about the two feature removal strategies.
Section~\ref{sec:appendix-batch-sage} contains Covert et al. \cite{Covert.2020}'s original batch SAGE algorithm and illustrates the pitfalls of applying it in an dynamic learning environment.
Finally, Section~\ref{sec:appendix-experiments} contains additional information about the data sets and streams, models and experimental setup used in Section~\ref{sec:experiments}.
Section~\ref{sec:appendix-experiments} also includes further experimental results.

\section{Proofs}
\label{sec:appendix-proofs}
All proofs are based on the static environment assumption, i.e. $f_t \equiv f$, $(X_t,Y_t) \equiv (X,Y)$ and $\mathbb{Q}^{(x,S)}_t$ is the true marginal (interventional) or conditional (observational) distribution with samples are drawn iid, similar to Covert et al. \cite{Covert.2020}.
\subsection{Proof of Theorem \ref{thm::consistency}}
\begin{proof}
With $M \to \infty$ and sampling iid from the true marginal distribution for the interventional approach and the conditional distribution for the observational approach, the law of large numbers ensures that the approximation converges, i.e. $\hat f_t(x,S) \to f_t(x,S)$ for every $t$ and $\emptyset \neq S \subset D$ almost surely.
Furthermore, for $S = \emptyset$, the smoothed mean prediction fulfills
\begin{equation*}
    \mathbb{E}[\hat f_t(x,\emptyset)] = \alpha \sum_{s=t_0}^t (1-\alpha)^{t-s} \mathbb{E}[f_s(x_s)] \overset{t \to \infty}{\to} \mathbb{E}_X[f(X)].
\end{equation*}
Hence,
\begin{align*}
    &\mathbb{E}_{\pi_t}[\mathbb{E}_{(X,Y)}[\Delta_t(i)]] =  \mathbb{E}_{\pi_t}[\mathbb{E}_{(X,Y)}[\ell(\hat f_t(x_t,u_i^-(\pi_t)),y_t)-\ell(\hat f_t(x_t,u_i^+(\pi_t)),y_t)]]
    \\
    &\overset{M\to\infty}{\rightarrow}\mathbb{E}_{\pi_t}[\nu(u_i^-(\pi_t))-\nu(u_i^+(\pi_t))] = \phi_t(i) = \phi(i),
\end{align*}
as $f \equiv f_t$.
Then, $\hat\phi_t(i)$ can be written as a weighted sum $\hat\phi_t(i) = \alpha \sum_{s=t_0}^t (1-\alpha)^{t-s} \Delta_s(i)$ and thus
\begin{align*}
\mathbb{E}[\hat\phi_t(i)] &= \alpha\sum_{s=t_0}^t (1-\alpha)^{t-s} \mathbb{E}[\Delta_s(i)]
\\
&\overset{M\to\infty}{\rightarrow} \alpha\sum_{s=t_0}^t (1-\alpha)^{t-s} \phi(i) = \phi(i)(1-(1-\alpha)^{t-t_0+1}) \overset{t\to \infty}{\rightarrow} \phi(i).
\end{align*}
\end{proof}

\subsection{Proof of Theorem \ref{thm::approx-quality}}

\begin{proof}
The variance of $\Delta_t(i)$ is constant and we denote $\sigma^2 := \mathbb{V}[\Delta_t(i)]$, where the variance is taken over the distribution of $(X,Y,\pi_t,\tilde X_1,\dots,\tilde X_M)$, where $\pi_t \sim \text{unif}(\mathfrak S_D)$ and $\tilde X_1,\dots,\tilde X_M \sim \mathbb{Q}^{(x,S)}_t$.
Furthermore, for two time steps $s,t$ the random variables $\Delta_s(i),\Delta_t(i)$ are independent.
Hence,
\begin{align*}
    \mathbb{V}[\hat\phi_t(i)] = \alpha^2 \sum_{s=t_0}^t(1-\alpha)^{2(t-s)}  \mathbb{V}[\Delta_t(i)] \leq \sigma^2 \frac{\alpha}{2-\alpha} = \mathcal O (\alpha),
\end{align*}
where we have used the geometric series for $(1-\alpha)^{2}$ as an upper bound.
\end{proof}

\subsection{Proof of Theorem \ref{thm::approx-SAGE}}
\begin{proof}
It was shown in \cite{Covert.2020} that the SAGE estimator using $t$ samples fulfills $\mathbb{V}[\hat\phi_t^{\text{(SAGE)}}] = \mathcal O(\frac 1 t)$ and thus
\begin{align*}
    \mathbb{P}\left( \vert \hat\phi_t(i) - \hat\phi_t^{\text{SAGE}}(i) \vert > \epsilon \right) 
    &\leq \mathbb{P}\left( \vert \hat\phi_t(i) - \phi_t(i) \vert + \vert \phi_t(i) - \hat\phi_t^{\text{SAGE}}(i) \vert > \epsilon \right) 
    \\
    &\leq \mathbb{P}\left( \vert \hat\phi_t(i) - \phi_t(i) \vert > \epsilon\right) + \mathbb{P}\left(\vert \phi_t(i) - \hat\phi_t^{\text{SAGE}}(i) \vert > \epsilon \right) 
    \\
    &=  \mathbb{P}\left( \vert \hat\phi_t(i) - \phi_t(i) \vert > \epsilon\right) + \mathcal O(\frac 1 t).
\end{align*}
Now, for $M \to \infty$, the expectation of $\phi_t(i)$ is given as
\begin{align*}
    \mathbb{E}[\phi_t(i)] &= \alpha\sum_{s=t_0}^t (1-\alpha)^{t-s} \mathbb{E}[\Delta_s(i)]  
    \\
    &\overset{M\to\infty}{\rightarrow} \alpha\sum_{s=t_0}^t (1-\alpha)^{t-s} \phi(i) = \phi(i)(1-(1-\alpha)^{t-t_0+1}).
\end{align*}
Hence, by Chebyshev's inequality and Theorem \ref{thm::approx-quality}
\begin{align*}
     \mathbb{P}\left( \vert \hat\phi_t(i) - \phi_t(i) \vert > \epsilon\right) &\leq \mathbb{P}\left( \vert \hat\phi_t(i) - \mathbb{E}[\hat\phi_t(i)]\vert +  \vert\mathbb{E}[\hat\phi_t(i)]-\phi_t(i) \vert > \epsilon\right) 
     \\
     &= \mathbb{P}\left( \vert \hat\phi_t(i) - \mathbb{E}[\hat\phi_t(i)]\vert +  \vert (1-\alpha)^{t-t_0+1}\phi(i) \vert >  \epsilon\right) 
    \\
     &= \mathbb{P}\left( \vert \hat\phi_t(i) - \mathbb{E}[\hat\phi_t(i)]\vert >  \epsilon -  \vert (1-\alpha)^{t-t_0+1}\phi(i) \vert\right) 
     \\
     &\leq \mathcal O(\mathbb{V}[\hat\phi_t(i)]) = \mathcal O(\alpha) = \mathcal O(\frac 1 t),
\end{align*}
where the assumption ensures that $\epsilon -  \vert (1-\alpha)^{t-t_0+1}\phi(i) \vert > 0$.
Finally,
\begin{equation*}
     \mathbb{P}\left( \vert \hat\phi_t(i) - \hat\phi_t^{\text{SAGE}}(i) \vert > \epsilon \right) = \mathcal O(\frac 1 t).
\end{equation*}
\end{proof}

\newpage

\section{Interventional and Observational Removal Strategies}
\label{sec:appendix-marginal-tree-feature-removal}

We propose two distinct feature removal strategies for the \emph{interventional} and the \emph{observational} iSAGE as described in Section~\ref{sec:background} and in particular Section~\ref{sec:feature_removal_algorithms}.
For the interventional approach, we propose to approximate the marginal feature distribution incrementally (see Appendix~\ref{sec:appendix-marginal-feature-removal}). 
For the observational approach we propose to approximate the conditional data distribution (see Appendix~\ref{sec:appendix-conditional-tree-feature-removal}).
Both approximation strategies can be efficiently computed and used to sample replacement values to restrict the model function for calculating the SAGE values.

\begin{figure*}[b]
    \centering
    \begin{minipage}[t]{0.32\textwidth}
        \includegraphics[width=\textwidth]{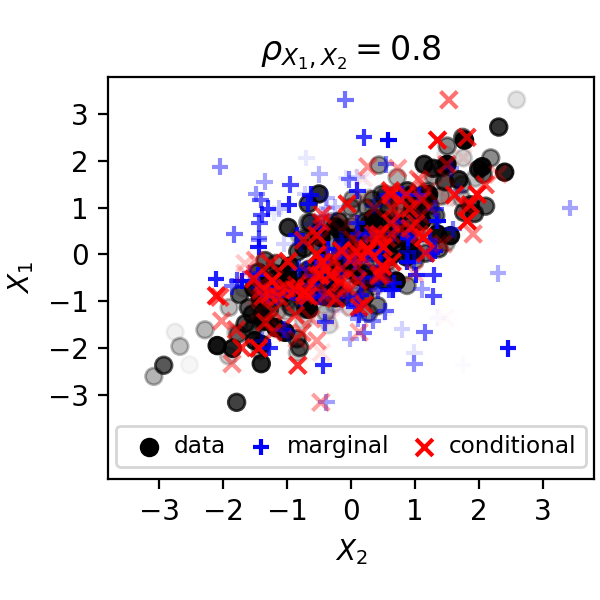}
    \end{minipage}
    \begin{minipage}[t]{0.65\textwidth}
        \includegraphics[width=\textwidth]{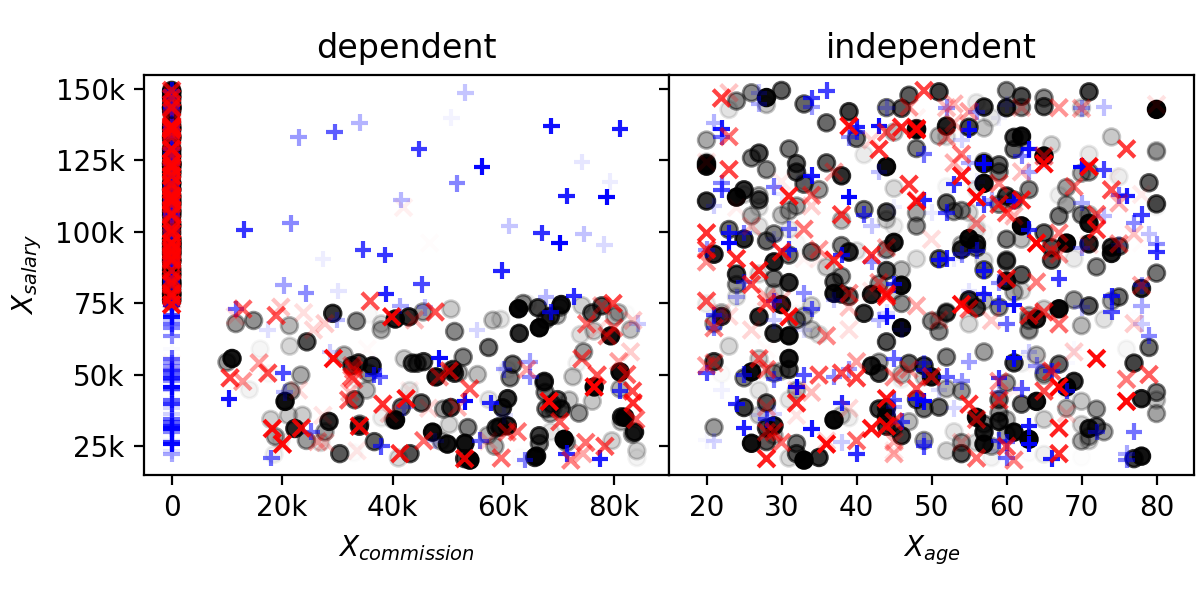}
    \end{minipage}
    \caption{Observational (red) and interventional (blue) feature removal strategies; The features follow the distributions $X_1 \sim \mathcal N(0, 1)$, $X_2 \sim \mathcal N(0, 1)$, $X_{\text{age}} \sim \text{unif}([20, 80])$, $X_{\text{salary}} \sim \text{unif}([20k, 150k])$, and $X_{\text{commission}} = \mathbf{1}(X_{\text{salary}}\leq 75k) \cdot Q$ with $Q \sim \text{unif}([10k,75k])$.}
    \label{fig:marginal-conditional-feature-removal}
\end{figure*}

Fig.~\ref{fig:marginal-conditional-feature-removal}, shows how both sampling approaches approximate the data distribution over time.
The observational sampling approach (depicted with red crosses) clearly adheres to the dependencies in the data distribution, whereas the interventional approach breaks these dependencies.
Also, in the setting without dependencies, both methods do not differ substantially as the observational approximation techniques reduces to approximating the interventional distribution.

\subsection{Marginal (Interventional) Feature Removal with Reservoir Sampling}
\label{sec:appendix-marginal-feature-removal}

As described in Section~\ref{sec:feature_removal_algorithms}, we apply geometric reservoir sampling to approximate a current estimate of the marginal feature distribution.
The procedure in Algorithm~\ref{alg:marginal-feature-removal-update} describes how new observations are stored in the reservoir.
Sampling from the constructed reservoir is trivially defined by uniformly drawing a stored data point.

\begin{algorithm}[!ht]
{\small
\caption{Updating the incremental geometric reservoir storage as described in \protect\cite{Fumagalli.2022}}
\label{alg:marginal-feature-removal}
\label{alg:marginal-feature-removal-update}
\begin{algorithmic}[1]
\REQUIRE stream $\left\{x^{D}_{t}\right\}_{t=1}^{\infty}$ feature indices $D \gets \left\{1, 2,\dots, d\right\}$
\STATE Initialize reservoir $R \gets \emptyset$ and number of seen samples with $n$
\FORALL{$x_{t} \in$ stream}
    \STATE $n \gets n + 1$
    \IF{$|R| \leq n$}
        \STATE $R \gets R \cup x_t$
    \ELSE
        \STATE  $x_{\text{del}} \gets \textsc{SampleUniformly}(R)$ \COMMENT{sample observation to remove from reservoir}
        \STATE  $R \gets (R \setminus \{x_{\text{del}}\}) \cup \{x_t\}$ \COMMENT{replace $x_{\text{del}}$ with $x_t$}
    \ENDIF
\ENDFOR
\end{algorithmic}
}
\end{algorithm}

\subsection{Conditional (Observational) Feature Removal with Incremental Decision Trees}
\label{sec:appendix-conditional-tree-feature-removal}
As described in Section~\ref{sec:feature_removal_algorithms}, we propose to train efficient, incremental decision tree models to more closely approximate the conditional data distribution.
The procedure to train these incremental decision trees is given in Algorithm~\ref{alg:incremental-tree-storage}.

\begin{algorithm}[!ht]
{\small
\caption{Updating the incremental trees storage mechanism for efficient conditional feature removal}
\label{alg:incremental-tree-storage}
\begin{algorithmic}[1]
\REQUIRE stream $\left\{x^{D}_{t}\right\}_{t=1}^{\infty}$ feature indices $D \gets \left\{1, 2,\dots, d\right\}$
\STATE Initialize the tree $h^{i}_0$, its leafs $L^{i}_0 \gets \emptyset$, the data reservoirs $R^{i}_0 \gets \emptyset$, and a leaf-reservoir mapping $M^{i}(.)$ for all $i \in D$
\FORALL{$x_{t} \in$ stream}
    \FORALL{$i \in D$}
    \STATE $y^{i}_{t} \gets x^{i}_t$
    \STATE $x^{r} \gets x^{D \setminus \{i\}}_t$ \COMMENT{take rest of $x$ as input features}
    \STATE $h^{i}_{t} \gets \text{\textsc{learn\_one}}\left(h^{i}_{t-1}, x^{r}, y^{i}_{t}\right)$ \COMMENT{makes one incremental learning step with the remaining features as input}
    \STATE $L^{i}_{t} \gets \text{\textsc{get\_leafs}}\left(h^{i}_{t}\right)$ \COMMENT{traverses the tree and collects all leaf nodes}
    \STATE $R^{i}_{t} \gets R^{i}_{t-1} \cup \text{\textsc{initialize}}\left(M^{i}\left(L^{i}_{t} \setminus L^{i}_{t-1}\right)\right)$ \COMMENT{initialize new reservoirs at new leaf nodes}
    \STATE $R^{i}_{t} \gets R^{i}_{t} \setminus M^{i}(L^{i}_{t-1} \setminus L^{i}_{t})$ \COMMENT{delete outdated reservoirs}
    \STATE $l_{t}^{i} \gets \text{\textsc{predict\_leaf}}\left(h_{t}^{i}, x^{D \setminus i}\right)$ \COMMENT{get the leaf node associated with a prediction given the remaining features}
    \STATE $r_{t-1}^{i} \gets M^{i}(l_{t}^{i})$ \COMMENT{get the reservoir associated with the leaf node}
    \STATE $r_{t}^{i} \gets r_{t-1}^{i} \cup x_t$ \COMMENT{update the reservoir with current sample}
    \ENDFOR
\ENDFOR
\STATE \textbf{return} all trees $\text{h}^{i}_t$ and reservoirs $R^{i}_t$ for $i \in D$
\end{algorithmic}
}
\end{algorithm}

\begin{algorithm}[!ht]
{\small
\caption{Leaf Traversal Procedure}
\label{alg:incremental-tree-traversal}
\begin{algorithmic}[1]
\STATE \text{\textbf{procedure} \textsc{Traverse}(node: $n$, sample: $x$, features present: $S$)}:
\STATE Initialize sampling ratios $W \gets \emptyset$
\IF{$n$ is split node}
    \IF{$n$ splits on a feature present in $S$}
        \STATE $c \gets \textsc{GetNextChild}\left(x\right)$ \COMMENT{make the split according to $x$}
        \STATE $n \gets \textsc{Traverse}\left(c, x\right)$
        \STATE \textbf{return} $n$
    \ENDIF
    \STATE $C \gets \textsc{GetChildren}\left(n\right)$ \COMMENT{get all children of $n$}
    \FORALL{nodes $c \in C$}
        \STATE $W \gets W \cup \textsc{GetWeight}\left(c\right)$ \COMMENT{get weight of child in terms of how many samples have visited}
    \ENDFOR
    \STATE $c \gets \textsc{SampleWithWeight}\left(C, W\right)$ \COMMENT{convert weights into probabilites and sample a child node accordingly}
    \STATE $n \gets \textsc{Traverse}\left(c, x\right)$
     \STATE \textbf{return} $n$
\ENDIF
\STATE \textbf{return} $n$ \COMMENT{node $n$ is a leaf node}
\end{algorithmic}
}
\end{algorithm}

\clearpage
\section{The Batch SAGE Algorithm}
\label{sec:appendix-batch-sage}

\begin{figure*}[!ht]
    \centering
    \includegraphics[width=0.99\textwidth]{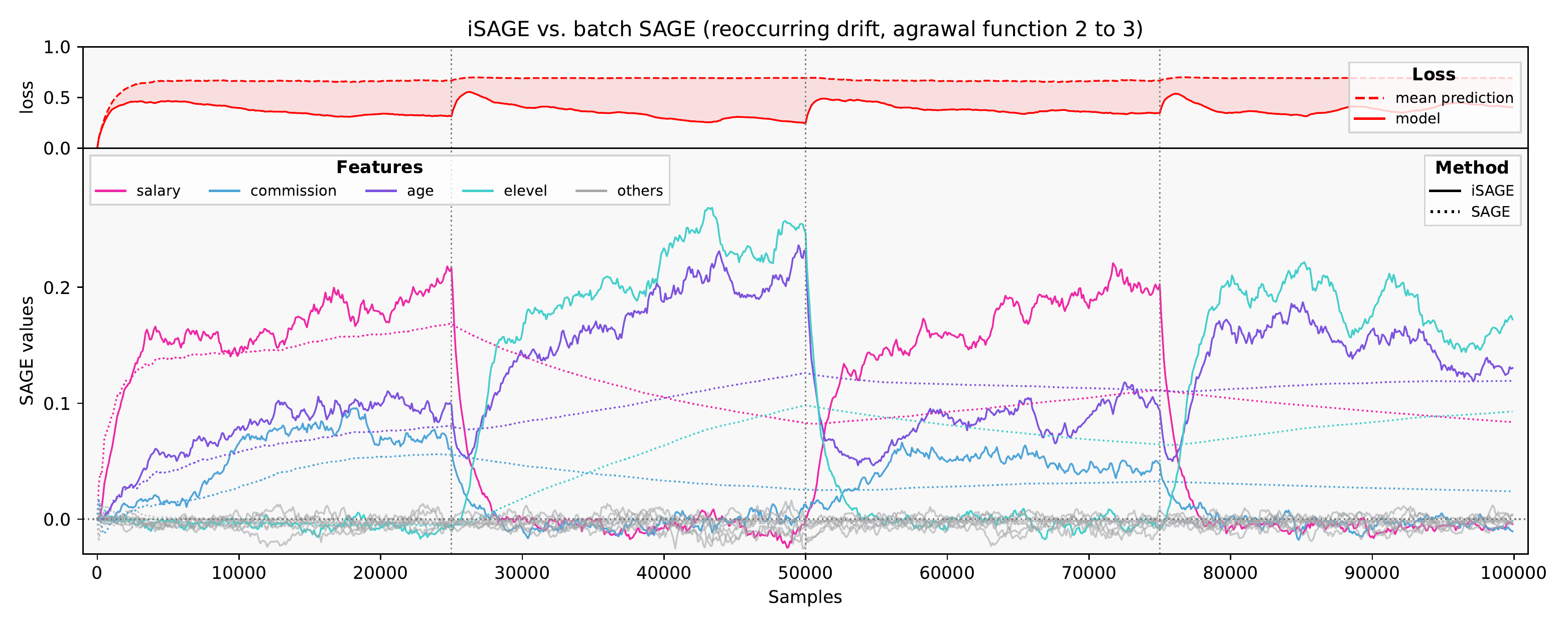}
    \caption{Static SAGE in Dynamic Learning Environment: The original SAGE (dash-dotted) is calculated in a dynamic, incremental learning scenario. Classical SAGE yields false importance scores when it is applied for a changing model, as it gives each importance score equal weight. iSAGE (solid) with $\alpha = 0.001$, and $m = 5$ is provided as a reference.}
    \label{fig:static_sage_in_dynamic_setting}
\end{figure*}

Algorithm~\ref{alg:original-sage} contains the original sampling-based SAGE algorithm by Covert et al. \cite{Covert.2020}.
The algorithm's notation is adjusted to fit into this paper's mathematical notation.
Fig.~\ref{fig:static_sage_in_dynamic_setting} illustrates the pitfall of naively applying SAGE (as defined in Algorithm~\ref{alg:original-sage}) in an incremental setting. 
The resulting importance values are incorrect because SAGE attributes equal weight onto each individual approximation step (in Lines 2 and 13 of Algorithm~\ref{alg:original-sage}).
Hence, older estimates that are no longer in-line with the real importance scores of a ever-changing model are given the same weight as more recent estimates.

\begin{algorithm}[!ht]
\caption{Sampling-based approximation for SAGE values \protect\cite{Covert.2020}}
\label{alg:original-sage}
\begin{algorithmic}[1]
\REQUIRE data $\left\{ x_{i}, y_{i}\right\}_{i=1}^{N}$, model $f$, loss function $l$, outer samples $n$, inner samples $m$ 
\STATE Initialize $\hat\phi^{1} \gets 0, \hat\phi^{2}\gets 0,\dots,\hat\phi^{d} \gets 0$
\STATE $y_\emptyset \gets \frac{1}{N} \sum_{i=1}^{N} f(x_i)$ \COMMENT{Marginal Prediction}
\FORALL{$(x_i,y_i) \in$ data}
    \STATE Sample $\pi$, a permutation of $D$
    \STATE $S \gets \emptyset$
    \STATE $ \text{lossPrev} \gets \ell( y_\emptyset,y_{i})$
    \FOR{$j = 1$ to $d$}
        \STATE $S \gets S \cup \left\{ \pi[j] \right\}$
        \STATE $y \gets 0$
        \FOR{$k = 1$ to $m$}
            \STATE Sample $x^{(\bar S)}_{k} \sim \mathbb Q^{(x,S)}$ \COMMENT{In practice: $\mathbb{P}(X^{(\bar S)})$}
            \STATE $y \gets y + f(x^{(S)}_{i}, x^{(\bar S)}_{k})$
        \ENDFOR
        \STATE $\bar y \gets \frac{y}{m}$
        \STATE $\text{loss} \gets l(\bar y,y^{i})$
        \STATE $\Delta \gets \text{lossPrev} - \text{loss}$
        \STATE $\hat \phi^{\pi[j]} \gets \phi^{\pi[j]} + \Delta$
        \STATE $\text{lossPrev} \gets \text{loss}$
    \ENDFOR
\ENDFOR
\STATE \textbf{return} $\frac{\phi^{1}}{n},\frac{\phi^{2}}{n},\dots,\frac{\phi^{d}}{n}$
\end{algorithmic}
\end{algorithm}

\clearpage

\section{Experiments}
\label{sec:appendix-experiments}

This section contains additional information about the conducted experiments.

\subsection{Data Set and Stream Descriptions}
\label{sec:appendix-data-set-description}

\paragraph{adult} 
Binary classification dataset that classifies $48842$ individuals based on 14 features into yearly salaries above and below 50k.
There are six numerical features and eight nominal features.
\emph{adult} is a publicly available dataset \cite{adult.1996}.

\paragraph{bank} 
Binary classification dataset that classifies $45211$ marketing phone calls based on 17 features to decide whether they decided to subscribe a term deposit.
There are seven numerical features and ten nominal features.
\emph{bank} is a publicly available dataset \cite{bank.2011}.

\paragraph{california}
Regression dataset containing $20640$ samples of 8 numerical features with 1990 census information from the US state of California. 
The dataset is publicly available at \url{https://www.dcc.fc.up.pt/~ltorgo/Regression/cal_housing.html}

\paragraph{bike} 
Regression dataset contains the hourly and daily count of rental bikes with information on weather. 
There are five numerical features and seven nominal features.
\emph{bike} is a publicly available dataset \cite{bike-batch.2014}.

\paragraph{agrawal} 
Synthetic data stream generator to create binary classification problems to decide whether an individual will be granted a loan based on nine features, six numerical and three nominal.
There are ten different decision functions available.
\emph{agrawal} is a publicly available dataset \cite{Agrawal.1993}.

\paragraph{stagger} 
The \emph{stagger} concepts makes a simple toy classification data stream. The synthetic data stream generator consists of three independent categorical features that describe the \emph{shape}, \emph{size}, and \emph{color} of an artificial object. Different classification functions can be derived from these sharp distinctions.
\emph{stagger} is a publicly available dataset \cite{stagger.1986}.

\paragraph{elec2} 
Binary classification dataset that classifies, if the electricity price will go up or down.
The data was collected for $45312$ time stamps from the Australian New South Wales Electricity Market and is based on eight features, six numerical and two nominal.
The data stream contains a well-documented concept drift in its \emph{vicprice} feature in that the feature has no values apart from zero in all observations up to $\approx20\,000$ samples.
After that the \emph{vicprice} feature starts having values different from zero.
\emph{elec2} is a publicly available dataset \cite{elec2.1999}.

\paragraph{L-Town Water Sensor Data}
L-Town is a popular variant of a virtual water distribution system, which is well-studied in the context of leakage detection algorithms \cite{Vrachimis.2022}.
Therein, sensor information can be simulated in different scenarios. 
We simulate pressure sensor readings over time and artificially introduce a sensor fault which needs to be detected.
The simulation tool is publicly available \cite{Klise.2017}.

\subsection{Summary of the Incremental Explanation Procedure}
\label{sec:appendix-incremental-learning-procedure}

Algorithm~\ref{alg:explanation-procedure} illustrates the simplified explanation procedure.
For each data point in a data stream, the incremental model first predicts the current's data point target label $y_t$ from $x_t$. 
This label is used for prequential evaluation of the model's performance and to calculate the model's loss at time $t$.
Then, the model is explained with this data point to update the the current iSAGE estimate $\hat{\phi}_t$.
After the explanation is updated the learning procedure is triggered for an incremental learning step.

\begin{algorithm}[!ht]
\caption{Incremental explanation procedure}
\label{alg:explanation-procedure}
\begin{algorithmic}[1]
\REQUIRE stream $\left\{ x_{t}, y_{t}\right\}_{t=1}^{\infty}$, model $f(.)$, loss function $\mathcal L(.)$
\FORALL{$(x_{t},y_{t}) \in$ stream}
    \STATE $\hat{y}_{t} \gets f_{t}(x_{t})$
    \STATE $\hat{\phi}_{t} \gets \text{\texttt{explain\_one}}(x_{t}, y_{t})$ 
    \STATE $f_{t + 1} \gets \text{\texttt{learn\_one}}(\mathcal L(\hat{y}_{t}, y_{t}))$
\ENDFOR
\end{algorithmic}
\end{algorithm}

\clearpage
\subsection{Further Experimental Results}
\label{app:further-experimental-results}

This subsection contains further results and details of the experiments conducted in Section~\ref{sec:experiments}.
First, we present additional results for the stationary data setting.
Second, we show examples of each synthetic GT scenario and provide further experimental results of different SW-SAGE window lengths and computational costs.
Third, we show additional examples of iSAGE in real incremental learning scenarios from different data streams.
Lastly, we provide the classification function to be learned omitted in Section~\ref{sec:experiments-feature-removal}.

\subsubsection{Approximation Quality in Static Batch Learning Scenarios}
\label{sec:appendix-batch}

\begin{figure}
    \centering
    \begin{minipage}{0.67\textwidth}
        \includegraphics[width=\textwidth]{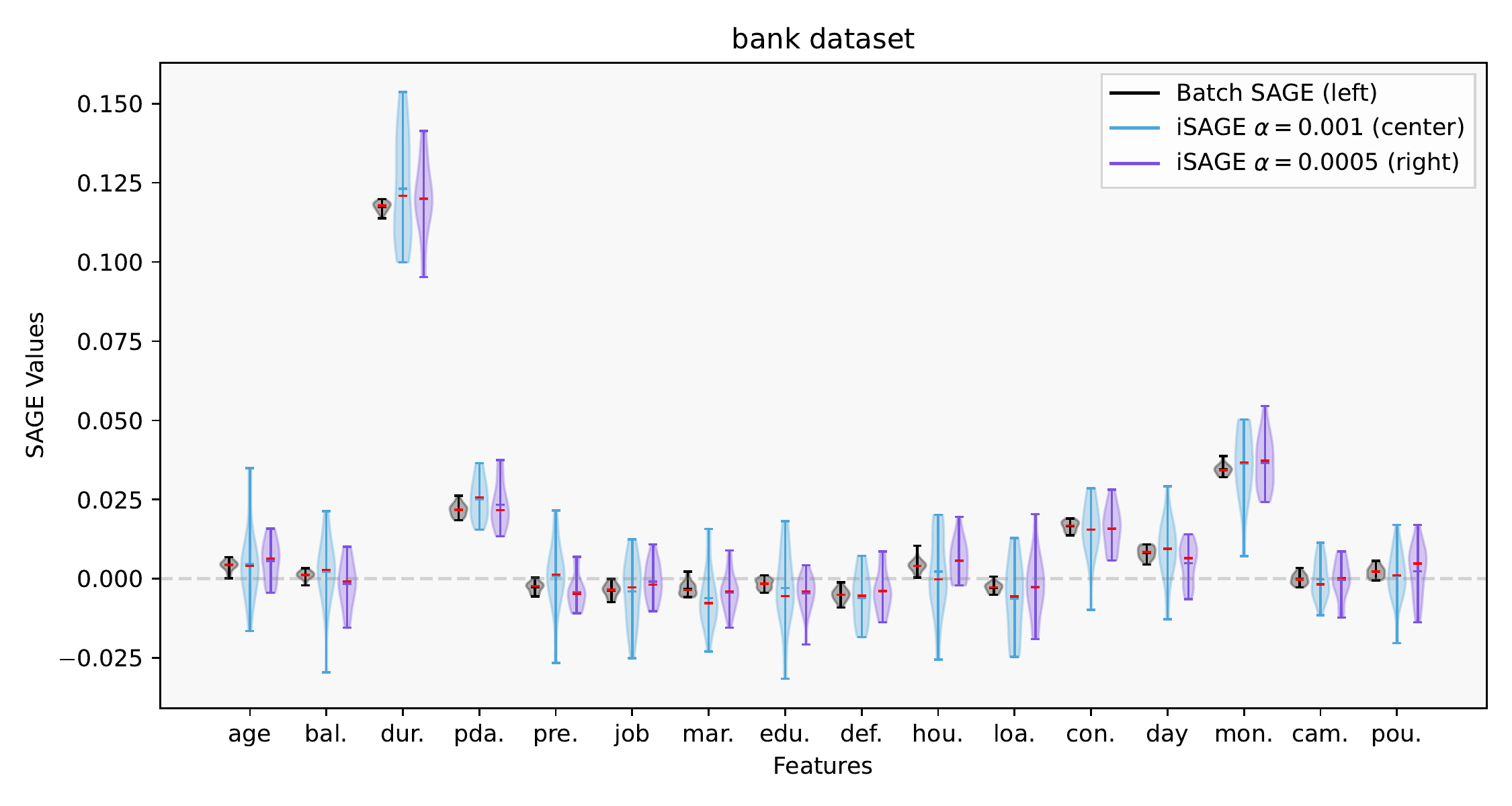}
    \end{minipage}
    \begin{minipage}{0.67\textwidth}
        \includegraphics[width=\textwidth]{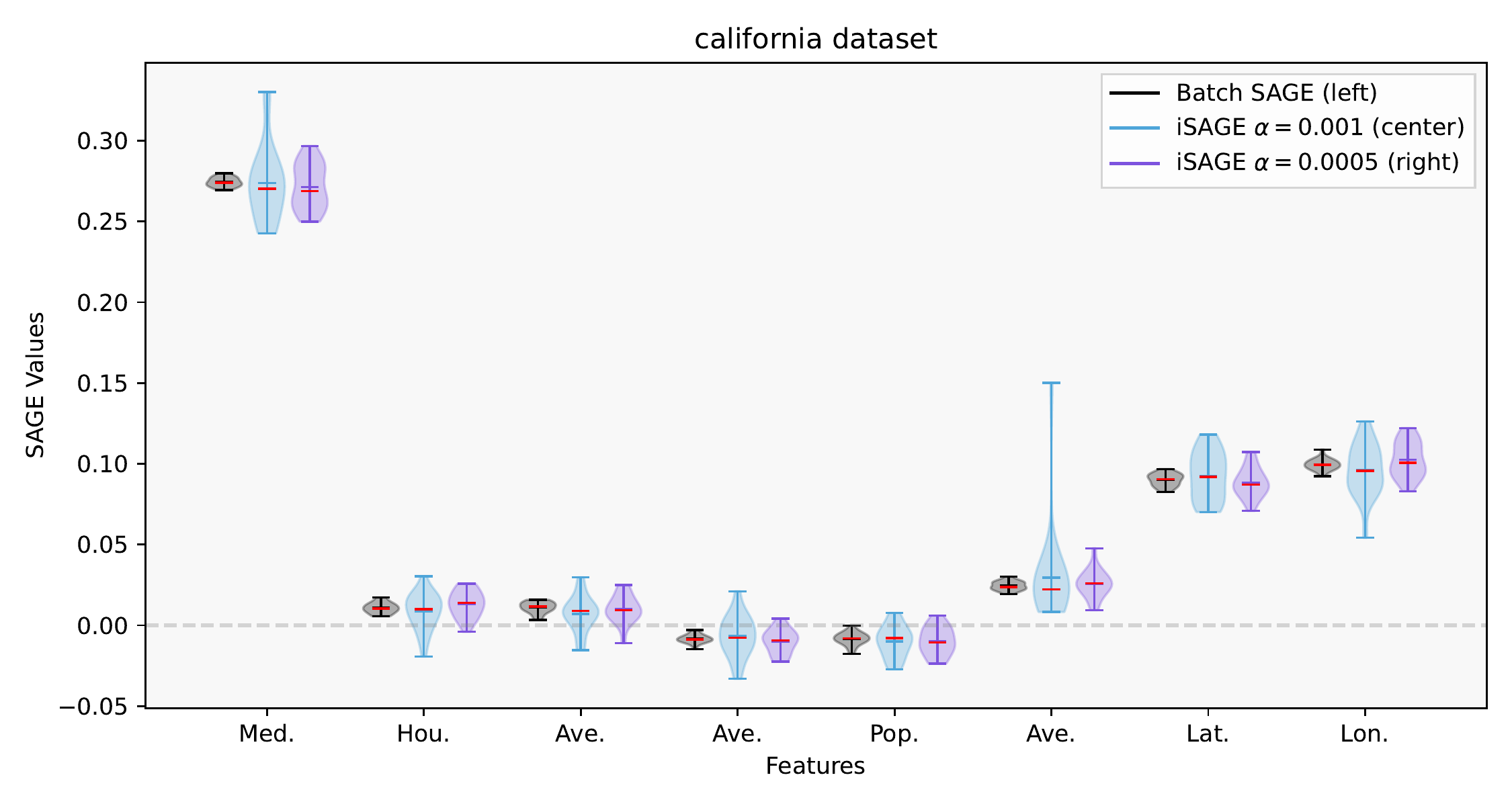}
    \end{minipage}
    \\
    \begin{minipage}{0.67\textwidth}
        \includegraphics[width=\textwidth]{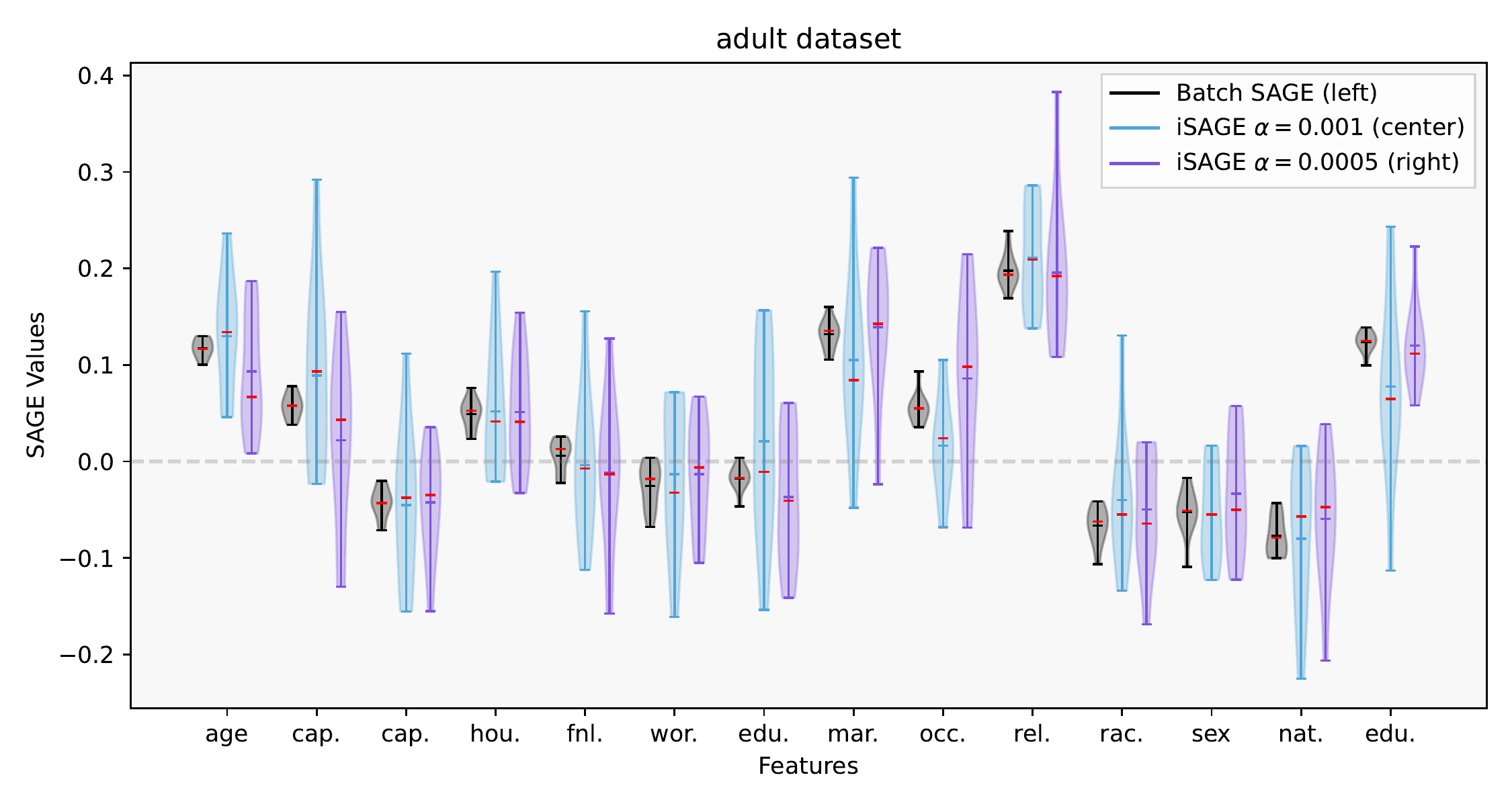}
    \end{minipage}
    \begin{minipage}{0.67\textwidth}
        \includegraphics[width=\textwidth]{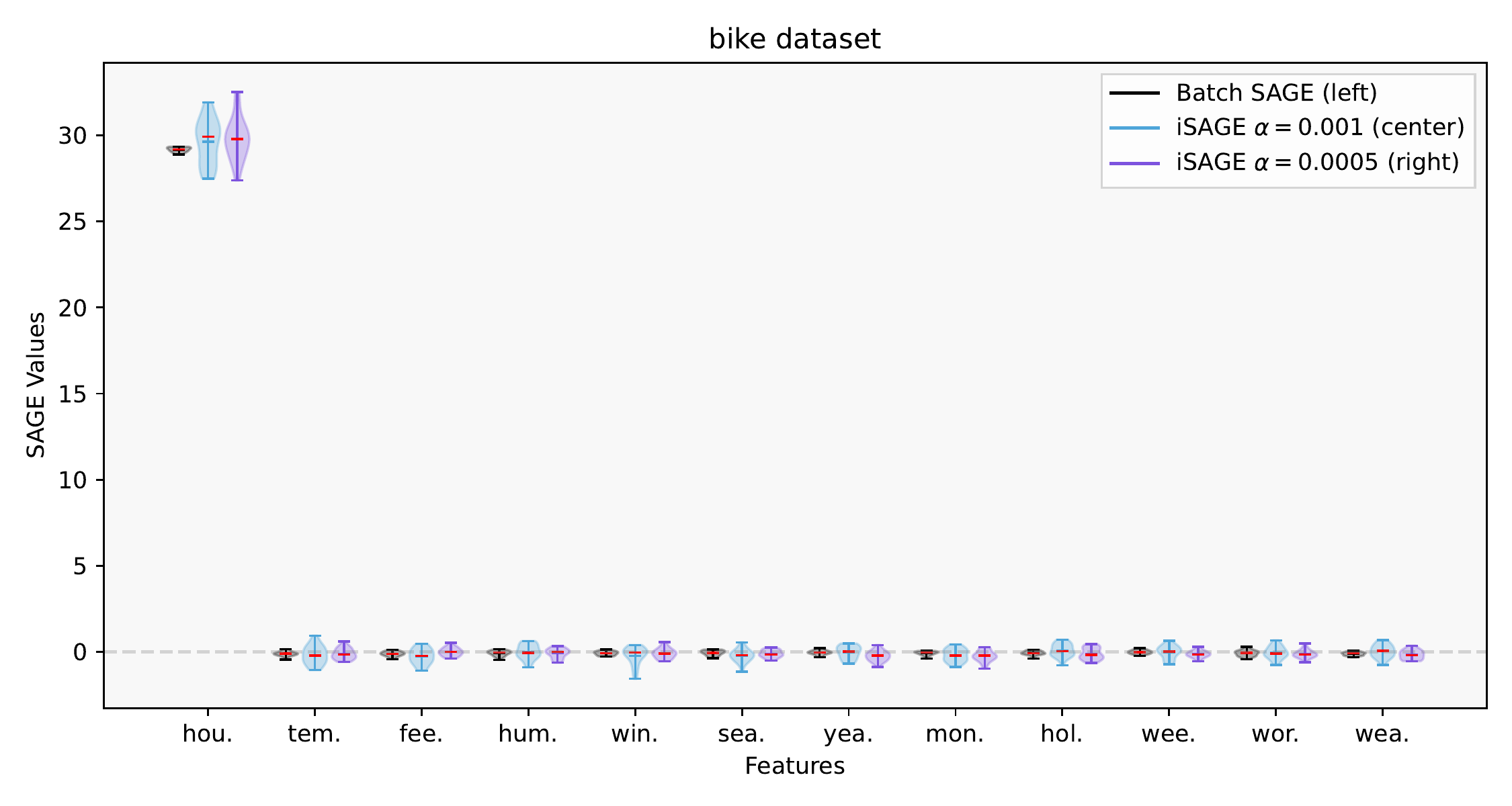}
    \end{minipage}
    \caption{SAGE values in stationary data setting calculated with batch SAGE and iSAGE for the \emph{bank} (1st row), \emph{california} (2nd row), \emph{adult} (3rd row), and \emph{bike} (4th row)  data sets.}
    \label{fig:batch_approximation}
\end{figure}

Fig.~\ref{fig:batch_approximation} contains SAGE values computed in a stationary data setting as described in \ref{sec:experiments-batch}. It shows the SAGE values for batch SAGE and iSAGE with $\alpha = 0.001$ and $\alpha = 0.0005$ for four datasets over mutliple runs.
For the \emph{bank} dataset, a LightGBM model was trained with \emph{max\_iterations} set to 70, a \emph{learning\_rate} of $0.2$ and a \emph{tree\_depth} of $15$. The LightGBM was explained in $20$ independent runs.
For the \emph{california} dataset, an ARF regressor was trained in an incremental manner with \emph{n\_models} set to 3 and explained in $20$ independent runs.
For the \emph{adult} dataset, an batch random forest classifier was trained with \emph{n\_models} set to 15 and explained in $10$ independent runs.
The \emph{bike} dataset, a HAT regressor was trained in an incremental manner and then explained in $20$ independent runs.

\subsubsection{Approximation Quality of Synthetic ground-truth Data Streams}
\label{sec:appendix-results-ground-truth}

\begin{figure}[t]
    \centering
    \includegraphics[width=0.99\textwidth]{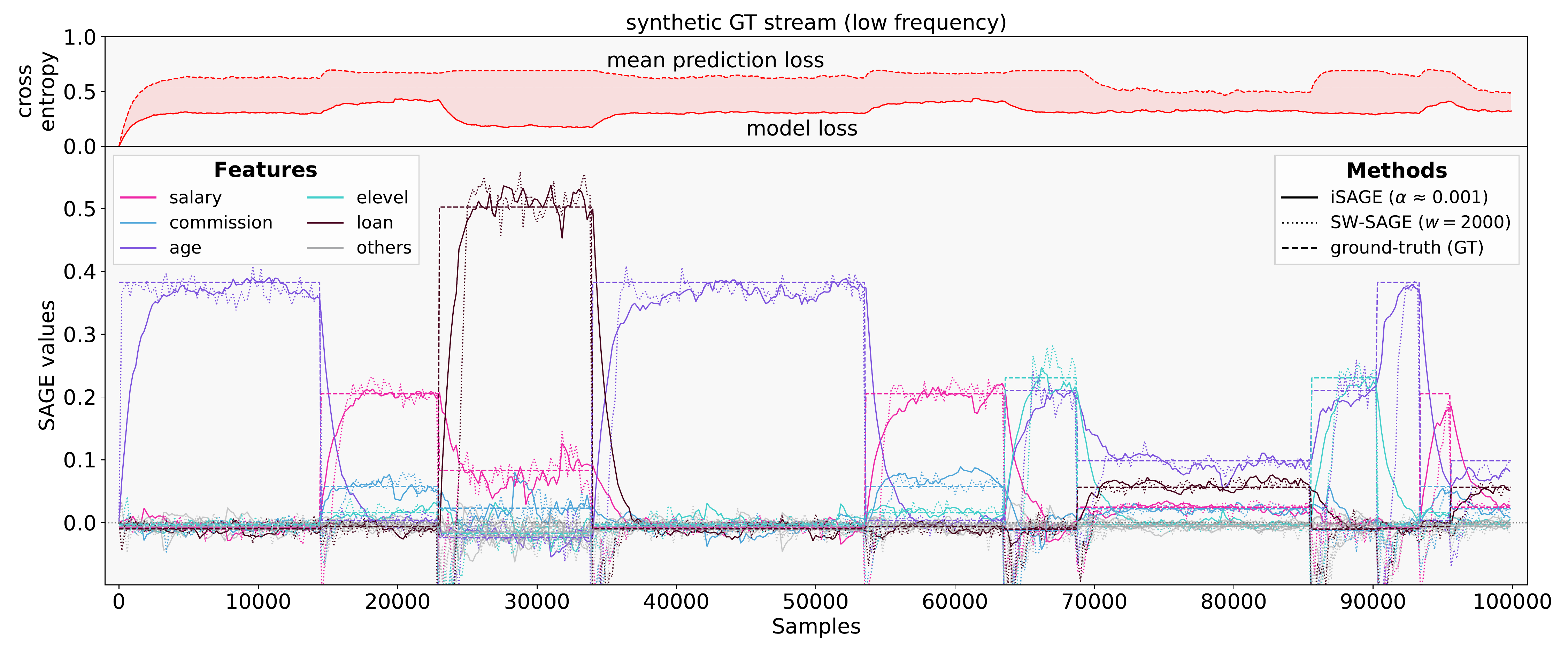}
    \includegraphics[width=0.99\textwidth]{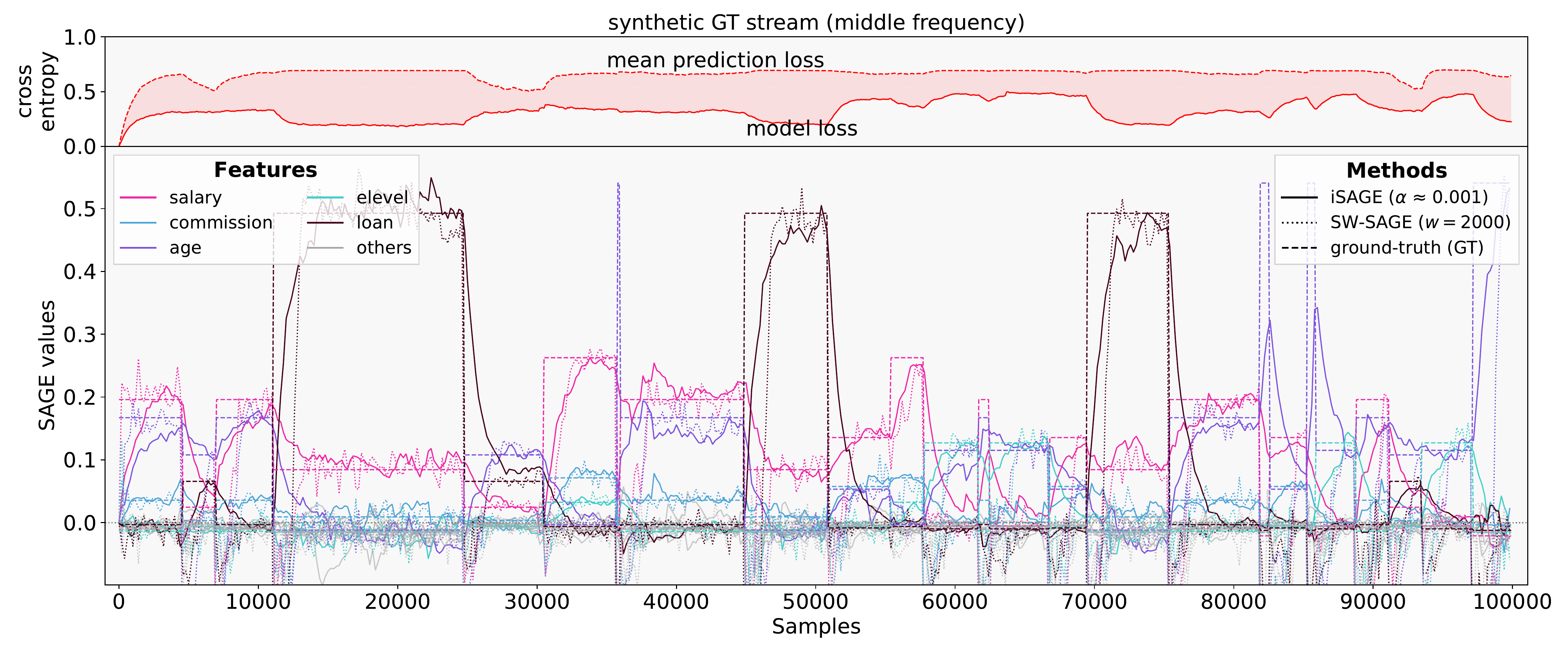}
    \includegraphics[width=0.99\textwidth]{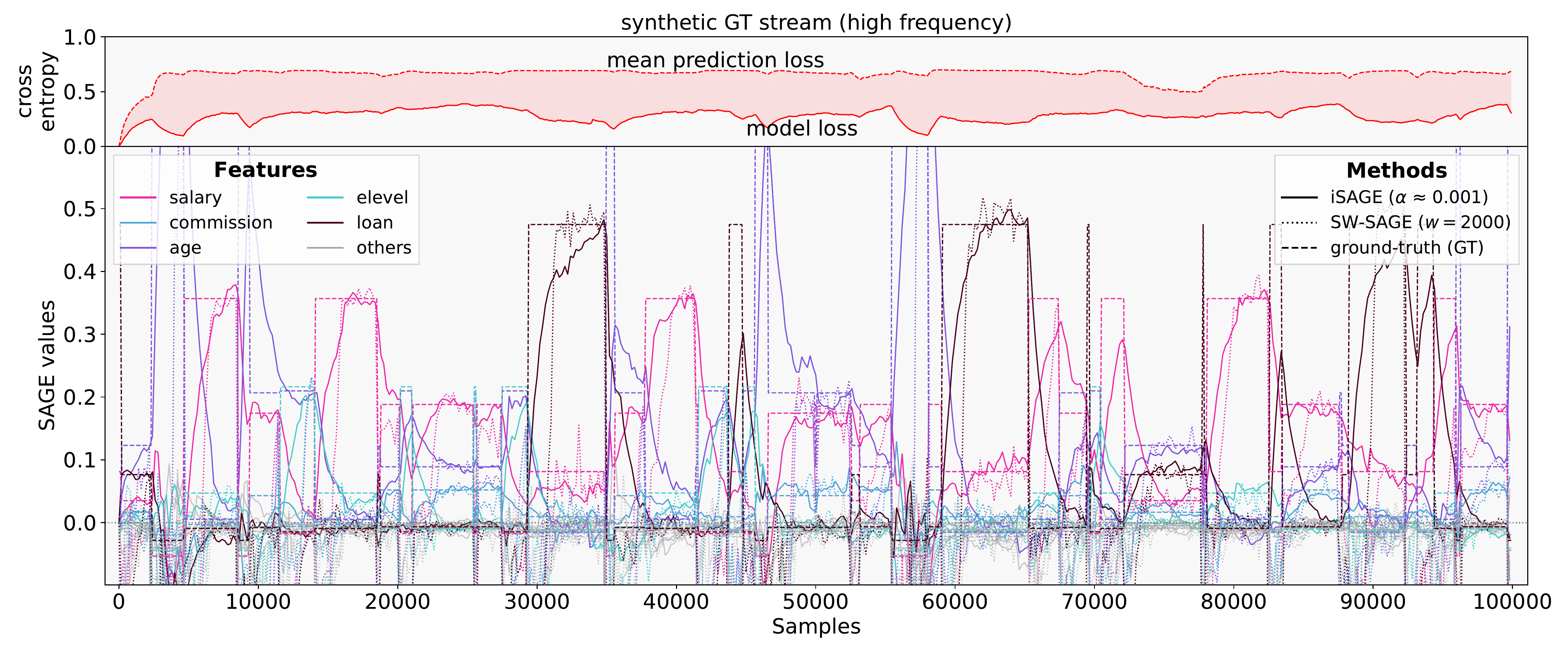}
    \caption{Three exemplary synthetic GT streams with iSAGE (solid), SW-SAGE (dotted), and the GT (dashed) for a low frequency (top), middle frequency (middle), and a high frequency (bottom) scenario. Presented are coupled iSAGE and SW-SAGE explanations with $alpha \approx 0.001$, and $w = 2000$ and $m = 4$, respectively.}
    \label{fig:appendix-gt-streams}
\end{figure}

In Fig.~\ref{fig:appendix-gt-streams} we present an exemplary data stream for each of the three scenarios of the synthetic GT experiments conducted in Section~\ref{sec:experiments-incremental}.
Fig.~\ref{fig:appendix-gt-streams} shows how iSAGE and SW-SAGE approximate the pre-computed GT SAGE values for the pre-trained models.
The runs in Fig.~\ref{fig:appendix-gt-streams} also show how SW-SAGE has a higher approximation error in times of change, whereas iSAGE gradually and smoothly switches between the concepts.
\\
For each scenario and iteration run, we pre-train incremental models, pre-compute the SAGE values in a batch mode, randomly shuffle the models in a synthetic data stream.
For each run, we pre-train six individual ARF classifiers (an ensemble with 3 HATs) on data generators based on the first six \emph{agrawal} concepts \cite{Agrawal.1993}.
We train each ARF for $20\,000$ samples.
After the pre-training, we compute the GT SAGE values according to Covert et al. \cite{Covert.2020}'s original SAGE definition.
Therein, we apply feature-removal according to the marginal feature distribution with $m = 10$.
Then, we create an artificial GT data stream by randomly switching between the different \emph{agrawal} data generators yielding different data stream distributions.
In each scenario, the probability of switching between the different pre-trained models is varied.
For the setting with high-, middle, and low-frequency of changes, we set the probability of switching at each time (sample point) to $p_{\text{switch}} = 0.0005 $, $p_{\text{switch}} = 0.0002$, and $p_{\text{switch}} = 0.0001$ respectively.
On average, these probabilities result in a model change after $2\,000$, $5\,000$, and $10\,000$ samples for the high-, middle, and low-frequency scenarios.
\\
In each scenario, we explain the underlying models with four different iSAGE and SW variants.
The SW span $w_1 = 100$, $w_2 = 500$, $w_3 = 1\,000$, and $w_4 = 2\,000$ samples, and are computed with a stride (frequency) of $w_i / 20$ samples.
We couple iSAGE's $\alpha$ parameter with the window sizes $\alpha_i = 2 / (w_i + 1)$ \cite{nahmias.2015}.
For both the SW-SAGE and iSAGE we set $m = 1$, the loss function to cross-entropy and apply the interventional feature removal approach.

\begin{figure}[t]
    \centering
    \includegraphics[width=0.95\textwidth]{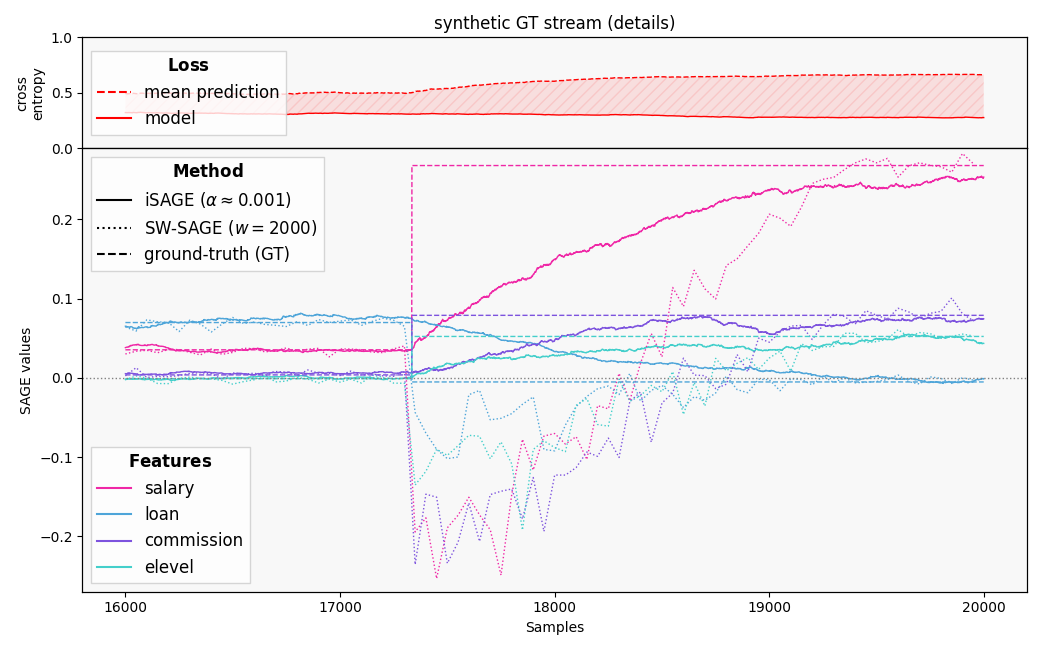}
    \caption{Detail view of a synthetic GT data stream. The models switch after $17\,335$ samples (different GT values). Before the switch, iSAGE and SW-SAGE approximate the GT well. Yet, after the switch, SW-SAGE recovers more slowly with a high approximation error.}
    \label{fig:appendix-gt-streams-detail-view}
\end{figure}

\begin{table}[t]
\centering
\caption{Approximation quality of iSAGE and SW-SAGE on synthetic GT data streams for $20$ iterations ($\text{inc}_{c}$ denotes iSAGE and $\text{SW}_{c}$ SW-SAGE with $c$ denoting the factor of additional computational cost compared to iSAGE). (std. in brackets)}
\label{tab:ground-truth-complete-results}
{
\setlength{\tabcolsep}{1pt}
\footnotesize
\begin{tabular}{@{}llcccc|cccc|cccc@{}}
\toprule
\multicolumn{2}{l}{\textbf{scenario}} & \multicolumn{4}{c}{\textbf{low}} & \multicolumn{4}{c}{\textbf{middle}} & \multicolumn{4}{c}{\textbf{high}} \\ \midrule
\multicolumn{2}{l}{\textbf{size ($w$)}} & $100$ & $500$ & $1\,000$ & $2\,000$ & $100$ & $500$ & $1\,000$ & $2\,000$ & $100$ & $500$ & $1\,000$ & $2\,000$ \\ \midrule
\multirow{6}{*}{\textbf{MAE}} & $\text{inc}_1$ & 
\begin{tabular}[c]{@{}c@{}}\textbf{.372}\\(.085)\end{tabular}&
\begin{tabular}[c]{@{}c@{}}\textbf{.197}\\(.052)\end{tabular}&
\begin{tabular}[c]{@{}c@{}}\textbf{.164}\\(.041)\end{tabular}&
\begin{tabular}[c]{@{}c@{}}\textbf{.153}\\(.040)\end{tabular}&
\begin{tabular}[c]{@{}c@{}}\textbf{.459}\\(.155)\end{tabular}&
\begin{tabular}[c]{@{}c@{}}\textbf{.254}\\(.080)\end{tabular}&
\begin{tabular}[c]{@{}c@{}}\textbf{.225}\\(.077)\end{tabular}&
\begin{tabular}[c]{@{}c@{}}\textbf{.226}\\(.074)\end{tabular}&
\begin{tabular}[c]{@{}c@{}}\textbf{.454}\\(.126)\end{tabular}&
\begin{tabular}[c]{@{}c@{}}\textbf{.284}\\(.075)\end{tabular}&
\begin{tabular}[c]{@{}c@{}}\textbf{.275}\\(.067)\end{tabular}&
\begin{tabular}[c]{@{}c@{}}\textbf{.305}\\(.072)\end{tabular} \\
 & $\text{SW}_{20}$ &
\begin{tabular}[c]{@{}c@{}}.384\\(.092)\end{tabular}&
\begin{tabular}[c]{@{}c@{}}.227\\(.057)\end{tabular}&
\begin{tabular}[c]{@{}c@{}}.219\\(.060)\end{tabular}&
\begin{tabular}[c]{@{}c@{}}.261\\(.085)\end{tabular}&
\begin{tabular}[c]{@{}c@{}}.492\\(.172)\end{tabular}&
\begin{tabular}[c]{@{}c@{}}.335\\(.111)\end{tabular}&
\begin{tabular}[c]{@{}c@{}}.361\\(.121)\end{tabular}&
\begin{tabular}[c]{@{}c@{}}.480\\(.175)\end{tabular}&
\begin{tabular}[c]{@{}c@{}}.505\\(.139)\end{tabular}&
\begin{tabular}[c]{@{}c@{}}.452\\(.130)\end{tabular}&
\begin{tabular}[c]{@{}c@{}}.575\\(.176)\end{tabular}&
\begin{tabular}[c]{@{}c@{}}.811\\(.249)\end{tabular} \\
 & $\text{SW}_{10}$ &
\begin{tabular}[c]{@{}c@{}}.384\\(.092)\end{tabular}&
\begin{tabular}[c]{@{}c@{}}.228\\(.057)\end{tabular}&
\begin{tabular}[c]{@{}c@{}}.220\\(.061)\end{tabular}&
\begin{tabular}[c]{@{}c@{}}.263\\(.084)\end{tabular}&
\begin{tabular}[c]{@{}c@{}}.492\\(.172)\end{tabular}&
\begin{tabular}[c]{@{}c@{}}.336\\(.111)\end{tabular}&
\begin{tabular}[c]{@{}c@{}}.365\\(.124)\end{tabular}&
\begin{tabular}[c]{@{}c@{}}.487\\(.180)\end{tabular}&
\begin{tabular}[c]{@{}c@{}}.506\\(.139)\end{tabular}&
\begin{tabular}[c]{@{}c@{}}.456\\(.131)\end{tabular}&
\begin{tabular}[c]{@{}c@{}}.580\\(.176)\end{tabular}&
\begin{tabular}[c]{@{}c@{}}.820\\(.254)\end{tabular} \\
 & $\text{SW}_5$ &
\begin{tabular}[c]{@{}c@{}}.384\\(.092)\end{tabular}&
\begin{tabular}[c]{@{}c@{}}.230\\(.058)\end{tabular}&
\begin{tabular}[c]{@{}c@{}}.221\\(.061)\end{tabular}&
\begin{tabular}[c]{@{}c@{}}.264\\(.081)\end{tabular}&
\begin{tabular}[c]{@{}c@{}}.494\\(.173)\end{tabular}&
\begin{tabular}[c]{@{}c@{}}.338\\(.112)\end{tabular}&
\begin{tabular}[c]{@{}c@{}}.371\\(.130)\end{tabular}&
\begin{tabular}[c]{@{}c@{}}.494\\(.184)\end{tabular}&
\begin{tabular}[c]{@{}c@{}}.508\\(.140)\end{tabular}&
\begin{tabular}[c]{@{}c@{}}.462\\(.132)\end{tabular}&
\begin{tabular}[c]{@{}c@{}}.592\\(.184)\end{tabular}&
\begin{tabular}[c]{@{}c@{}}.839\\(.257)\end{tabular} \\
 & $\text{SW}_2$ &
\begin{tabular}[c]{@{}c@{}}.385\\(.093)\end{tabular}&
\begin{tabular}[c]{@{}c@{}}.233\\(.061)\end{tabular}&
\begin{tabular}[c]{@{}c@{}}.235\\(.066)\end{tabular}&
\begin{tabular}[c]{@{}c@{}}.285\\(.092)\end{tabular}&
\begin{tabular}[c]{@{}c@{}}.494\\(.171)\end{tabular}&
\begin{tabular}[c]{@{}c@{}}.349\\(.113)\end{tabular}&
\begin{tabular}[c]{@{}c@{}}.389\\(.133)\end{tabular}&
\begin{tabular}[c]{@{}c@{}}.534\\(.208)\end{tabular}&
\begin{tabular}[c]{@{}c@{}}.512\\(.142)\end{tabular}&
\begin{tabular}[c]{@{}c@{}}.482\\(.137)\end{tabular}&
\begin{tabular}[c]{@{}c@{}}.615\\(.190)\end{tabular}&
\begin{tabular}[c]{@{}c@{}}.879\\(.274)\end{tabular} \\
 & $\text{SW}_1$ &
\begin{tabular}[c]{@{}c@{}}.386\\(.092)\end{tabular}&
\begin{tabular}[c]{@{}c@{}}.244\\(.064)\end{tabular}&
\begin{tabular}[c]{@{}c@{}}.246\\(.070)\end{tabular}&
\begin{tabular}[c]{@{}c@{}}.306\\(.098)\end{tabular}&
\begin{tabular}[c]{@{}c@{}}.495\\(.171)\end{tabular}&
\begin{tabular}[c]{@{}c@{}}.364\\(.121)\end{tabular}&
\begin{tabular}[c]{@{}c@{}}.426\\(.160)\end{tabular}&
\begin{tabular}[c]{@{}c@{}}.566\\(.216)\end{tabular}&
\begin{tabular}[c]{@{}c@{}}.519\\(.143)\end{tabular}&
\begin{tabular}[c]{@{}c@{}}.506\\(.148)\end{tabular}&
\begin{tabular}[c]{@{}c@{}}.670\\(.214)\end{tabular}&
\begin{tabular}[c]{@{}c@{}}.921\\(.345)\end{tabular} \\ \midrule
\multirow{6}{*}{\textbf{MSE}} & $\text{inc}_1$ &
\begin{tabular}[c]{@{}c@{}}\textbf{.057}\\(.042)\end{tabular}&
\begin{tabular}[c]{@{}c@{}}\textbf{.015}\\(.012)\end{tabular}&
\begin{tabular}[c]{@{}c@{}}\textbf{.013}\\(.009)\end{tabular}&
\begin{tabular}[c]{@{}c@{}}\textbf{.015}\\(.011)\end{tabular}&
\begin{tabular}[c]{@{}c@{}}\textbf{.103}\\(.105)\end{tabular}&
\begin{tabular}[c]{@{}c@{}}\textbf{.027}\\(.023)\end{tabular}&
\begin{tabular}[c]{@{}c@{}}\textbf{.027}\\(.026)\end{tabular}&
\begin{tabular}[c]{@{}c@{}}\textbf{.034}\\(.034)\end{tabular}&
\begin{tabular}[c]{@{}c@{}}\textbf{.094}\\(.066)\end{tabular}&
\begin{tabular}[c]{@{}c@{}}\textbf{.034}\\(.021)\end{tabular}&
\begin{tabular}[c]{@{}c@{}}\textbf{.038}\\(.022)\end{tabular}&
\begin{tabular}[c]{@{}c@{}}\textbf{.051}\\(.027)\end{tabular} \\
 & $\text{SW}_{20}$ &
\begin{tabular}[c]{@{}c@{}}.066\\(.050)\end{tabular}&
\begin{tabular}[c]{@{}c@{}}.049\\(.043)\end{tabular}&
\begin{tabular}[c]{@{}c@{}}.078\\(.081)\end{tabular}&
\begin{tabular}[c]{@{}c@{}}.139\\(.150)\end{tabular}&
\begin{tabular}[c]{@{}c@{}}.139\\(.140)\end{tabular}&
\begin{tabular}[c]{@{}c@{}}.191\\(.271)\end{tabular}&
\begin{tabular}[c]{@{}c@{}}.320\\(.487)\end{tabular}&
\begin{tabular}[c]{@{}c@{}}.582\\(.971)\end{tabular}&
\begin{tabular}[c]{@{}c@{}}.151\\(.104)\end{tabular}&
\begin{tabular}[c]{@{}c@{}}.248\\(.198)\end{tabular}&
\begin{tabular}[c]{@{}c@{}}.420\\(.360)\end{tabular}&
\begin{tabular}[c]{@{}c@{}}.690\\(.607)\end{tabular} \\
 & $\text{SW}_{10}$ &
\begin{tabular}[c]{@{}c@{}}.065\\(.050)\end{tabular}&
\begin{tabular}[c]{@{}c@{}}.049\\(.044)\end{tabular}&
\begin{tabular}[c]{@{}c@{}}.080\\(.086)\end{tabular}&
\begin{tabular}[c]{@{}c@{}}.137\\(.148)\end{tabular}&
\begin{tabular}[c]{@{}c@{}}.139\\(.139)\end{tabular}&
\begin{tabular}[c]{@{}c@{}}.189\\(.266)\end{tabular}&
\begin{tabular}[c]{@{}c@{}}.325\\(.501)\end{tabular}&
\begin{tabular}[c]{@{}c@{}}.596\\(.035)\end{tabular}&
\begin{tabular}[c]{@{}c@{}}.152\\(.105)\end{tabular}&
\begin{tabular}[c]{@{}c@{}}.250\\(.198)\end{tabular}&
\begin{tabular}[c]{@{}c@{}}.422\\(.360)\end{tabular}&
\begin{tabular}[c]{@{}c@{}}.690\\(.612)\end{tabular} \\
 & $\text{SW}_5$ &
\begin{tabular}[c]{@{}c@{}}.066\\(.050)\end{tabular}&
\begin{tabular}[c]{@{}c@{}}.051\\(.047)\end{tabular}&
\begin{tabular}[c]{@{}c@{}}.078\\(.085)\end{tabular}&
\begin{tabular}[c]{@{}c@{}}.125\\(.125)\end{tabular}&
\begin{tabular}[c]{@{}c@{}}.141\\(.142)\end{tabular}&
\begin{tabular}[c]{@{}c@{}}.192\\(.273)\end{tabular}&
\begin{tabular}[c]{@{}c@{}}.337\\(.566)\end{tabular}&
\begin{tabular}[c]{@{}c@{}}.600\\(.062)\end{tabular}&
\begin{tabular}[c]{@{}c@{}}.155\\(.107)\end{tabular}&
\begin{tabular}[c]{@{}c@{}}.253\\(.199)\end{tabular}&
\begin{tabular}[c]{@{}c@{}}.429\\(.373)\end{tabular}&
\begin{tabular}[c]{@{}c@{}}.703\\(.628)\end{tabular} \\
 & $\text{SW}_2$ &
\begin{tabular}[c]{@{}c@{}}.065\\(.047)\end{tabular}&
\begin{tabular}[c]{@{}c@{}}.051\\(.048)\end{tabular}&
\begin{tabular}[c]{@{}c@{}}.088\\(.106)\end{tabular}&
\begin{tabular}[c]{@{}c@{}}.137\\(.145)\end{tabular}&
\begin{tabular}[c]{@{}c@{}}.137\\(.133)\end{tabular}&
\begin{tabular}[c]{@{}c@{}}.176\\(.226)\end{tabular}&
\begin{tabular}[c]{@{}c@{}}.326\\(.466)\end{tabular}&
\begin{tabular}[c]{@{}c@{}}.652\\(.249)\end{tabular}&
\begin{tabular}[c]{@{}c@{}}.156\\(.104)\end{tabular}&
\begin{tabular}[c]{@{}c@{}}.259\\(.211)\end{tabular}&
\begin{tabular}[c]{@{}c@{}}.432\\(.399)\end{tabular}&
\begin{tabular}[c]{@{}c@{}}.713\\(.644)\end{tabular} \\
 & $\text{SW}_1$ &
\begin{tabular}[c]{@{}c@{}}.066\\(.048)\end{tabular}&
\begin{tabular}[c]{@{}c@{}}.061\\(.067)\end{tabular}&
\begin{tabular}[c]{@{}c@{}}.080\\(.079)\end{tabular}&
\begin{tabular}[c]{@{}c@{}}.160\\(.215)\end{tabular}&
\begin{tabular}[c]{@{}c@{}}.136\\(.125)\end{tabular}&
\begin{tabular}[c]{@{}c@{}}.183\\(.200)\end{tabular}&
\begin{tabular}[c]{@{}c@{}}.399\\(.792)\end{tabular}&
\begin{tabular}[c]{@{}c@{}}.529\\(.883)\end{tabular}&
\begin{tabular}[c]{@{}c@{}}.162\\(.107)\end{tabular}&
\begin{tabular}[c]{@{}c@{}}.283\\(.262)\end{tabular}&
\begin{tabular}[c]{@{}c@{}}.462\\(.413)\end{tabular}&
\begin{tabular}[c]{@{}c@{}}.757\\(.891)\end{tabular} \\ \bottomrule
\end{tabular}
}
\end{table}

\subsubsection{Explaining Incremental Models}
\label{sec:appendix-incremental-models}

\begin{figure}[t]
    \centering
    \includegraphics[width=0.95\textwidth]{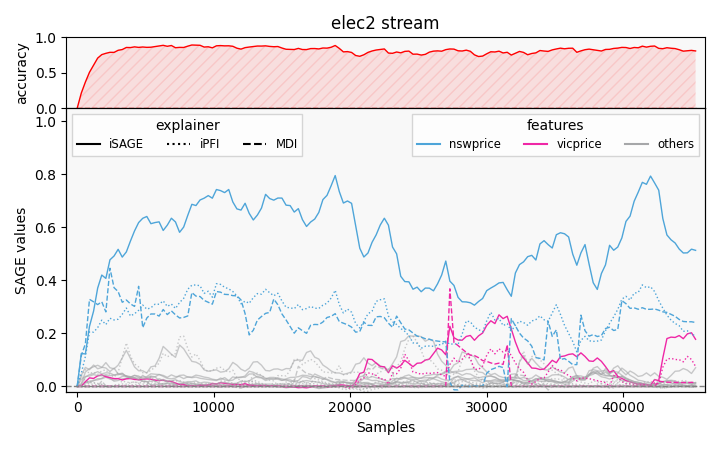}
    \caption{iSAGE, iPFI, and MDI for an HAT on \emph{elec2}}
    \label{fig:mdi_comparison}
    \vspace{-1em}
\end{figure}

Next to the experiments in Section~\ref{sec:experiments-incremental}, we compare iSAGE with related FI methods; namely, incremental permutation feature importance (iPFI) \cite{Fumagalli.2022} and mean decrease in impurity (MDI) \cite{Gomes.2019}.
MDI, as a model-specific method, is only applicable on incremental decision trees. 
Hence, in Fig.~\ref{fig:mdi_comparison}, we compute the three methods for a HAT on the \emph{elec2} \cite{elec2.1999} real world data stream as an example.
All methods correctly identify the two most important features.

\begin{figure}
    \centering
    \includegraphics[width=0.95\textwidth]{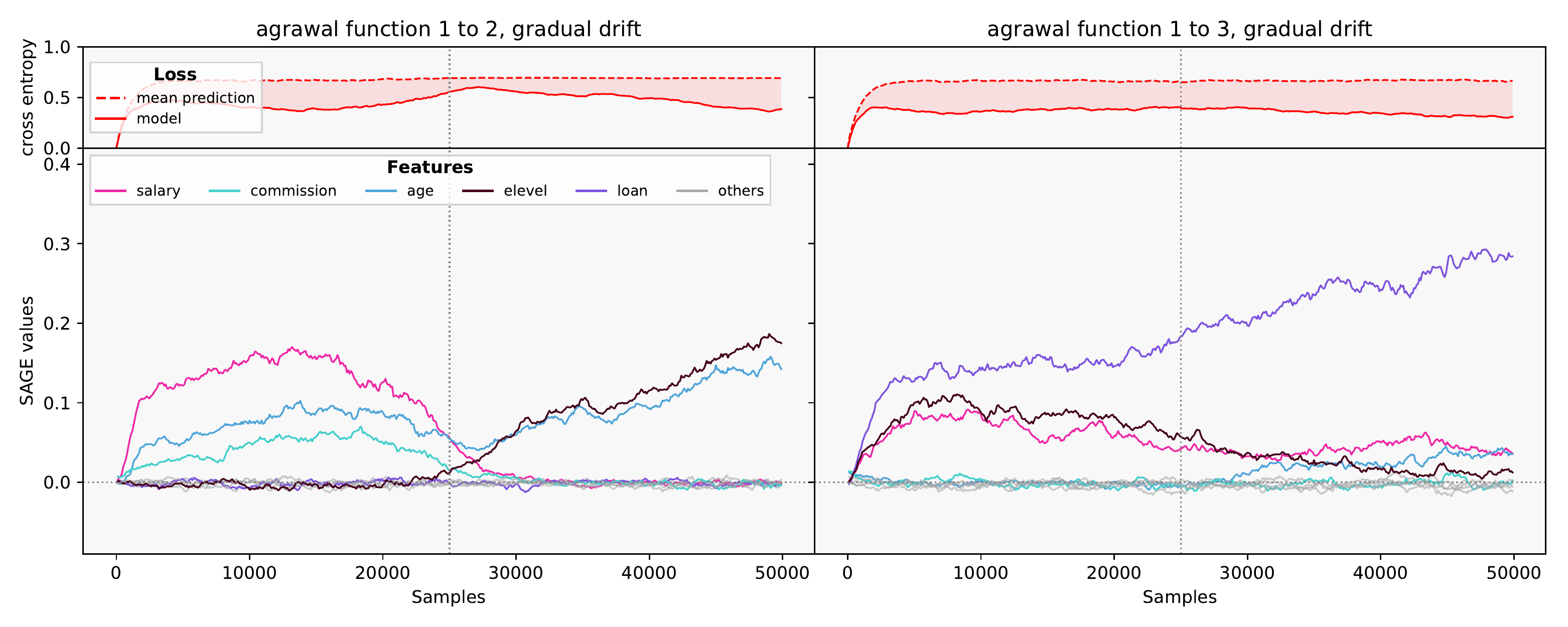}
    \includegraphics[width=0.95\textwidth]{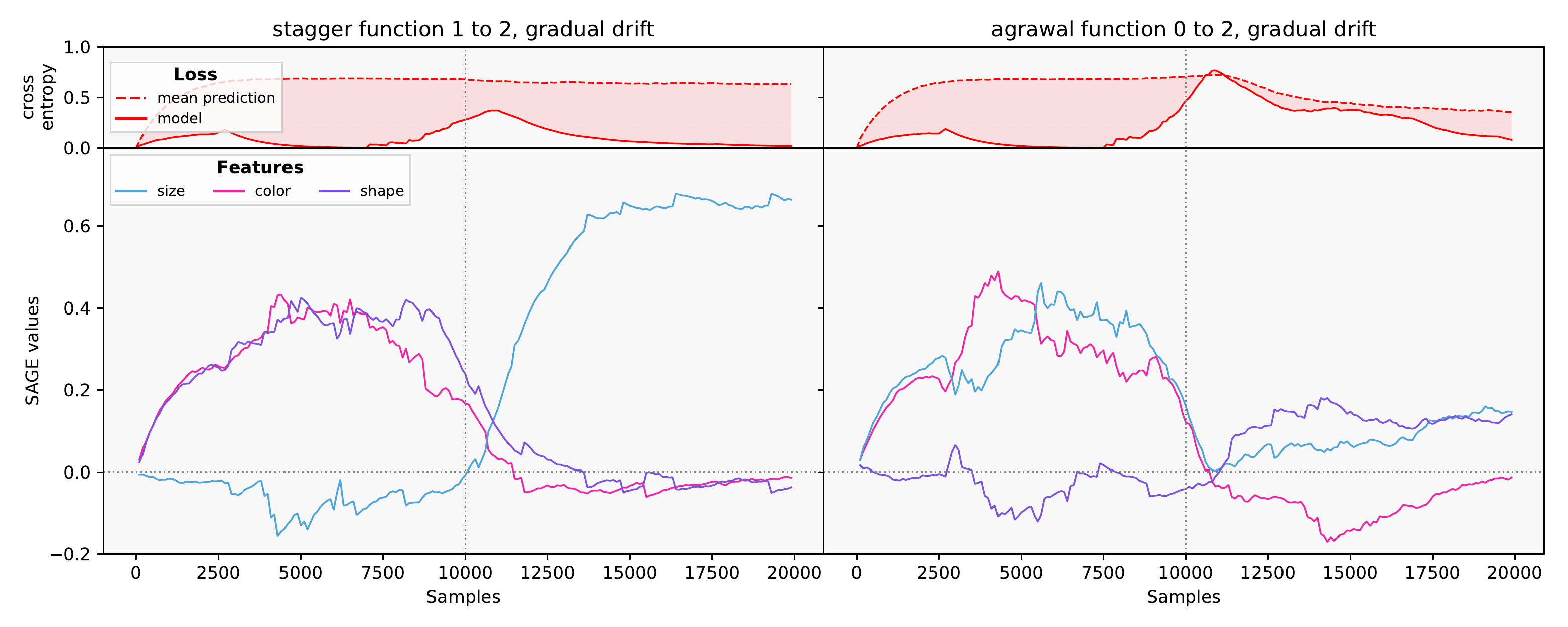}
    \caption{Two example cases of iSAGE explaining an ARF on an \emph{agrawal} gradual concept drift stream (top) and a HAT on a \emph{stagger} gardual concept drift stream (bottom)}
    \label{fig:examples_inc}
\end{figure}

\subsubsection{Online Sensor Fault Detection}
\label{sec:experiments_water_data}

\begin{figure}[t]
    \centering
    \includegraphics[width=0.5\textwidth]{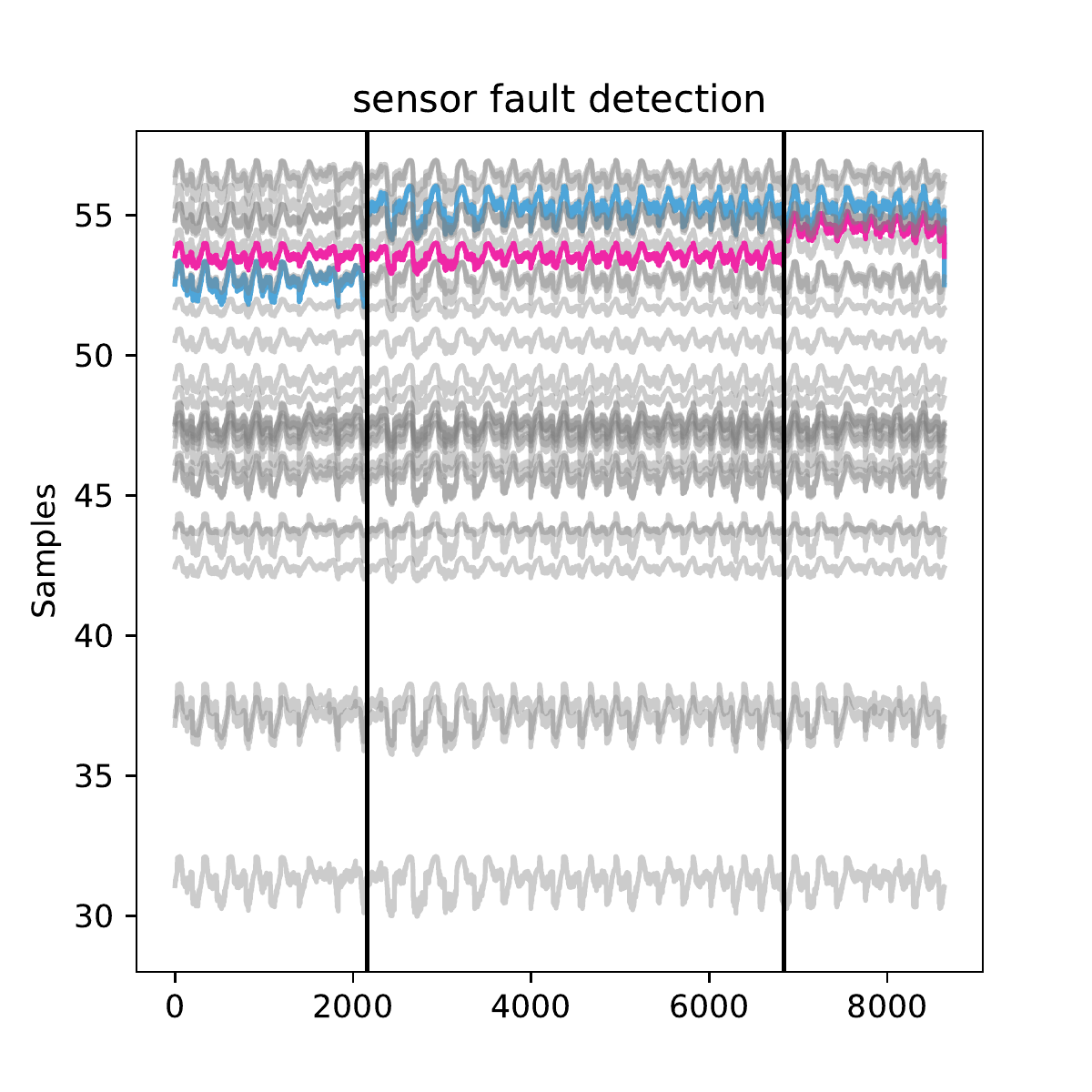}
    \caption{Sensor readings of the simulated online sensor network.}
    \label{fig:sensor_data}
\end{figure}

We conduct an experiment of online anomaly detection from sensor data.
Therein, we simulate a data stream of sensor readings using the L-Town virtual water distribution network \cite{Vrachimis.2022} with an open-source software \cite{Klise.2017} following Hinder et al. \cite{Hinder.2023}.
We create a data stream containing 29 pressure sensor readings (features).
Two sensor fault are randomly introduced (one at $t = 2160$ and one at $t = 6840$).
Fig.~\ref{fig:sensor_data} shows the dataset of sensor readings.
To illustrate how iSAGE can be used to explain any black-box model fitted on a data stream, we opted to fit an undercomplete autoencoder on the stream and explain its reconstruction error.
The autoencoder is trained with the \emph{river} and \emph{deep-river} \footnote{\url{https://github.com/online-ml/deep-river}} open source framework for training NN models incrementally. 
The autoencoder consists of 4 layers.
The encoder reduces the dimensionalty from the feature size to a hidden size of 10 and then to a latent dimension of 3.
The decoder reverses this by expanding the latent dimension of 3 to a second hidden size of 10 and then back to the feature dimension.
The input of the autoencoder is scaled with a standard scaler. 
The model is trained with a batch size of one (one model update with each new observation) with a relatively high learning rate of $\gamma = 0.05$. 
We explain this autoencoder with the intverventional iSAGE approach ($\alpha = 0.001$ and $m = 10$).

\subsubsection{Interventional and Observational iSAGE}
\label{sec:appendix-experiments-feature-removal}
\begin{figure}
    \centering
    \includegraphics[width=0.95\textwidth]{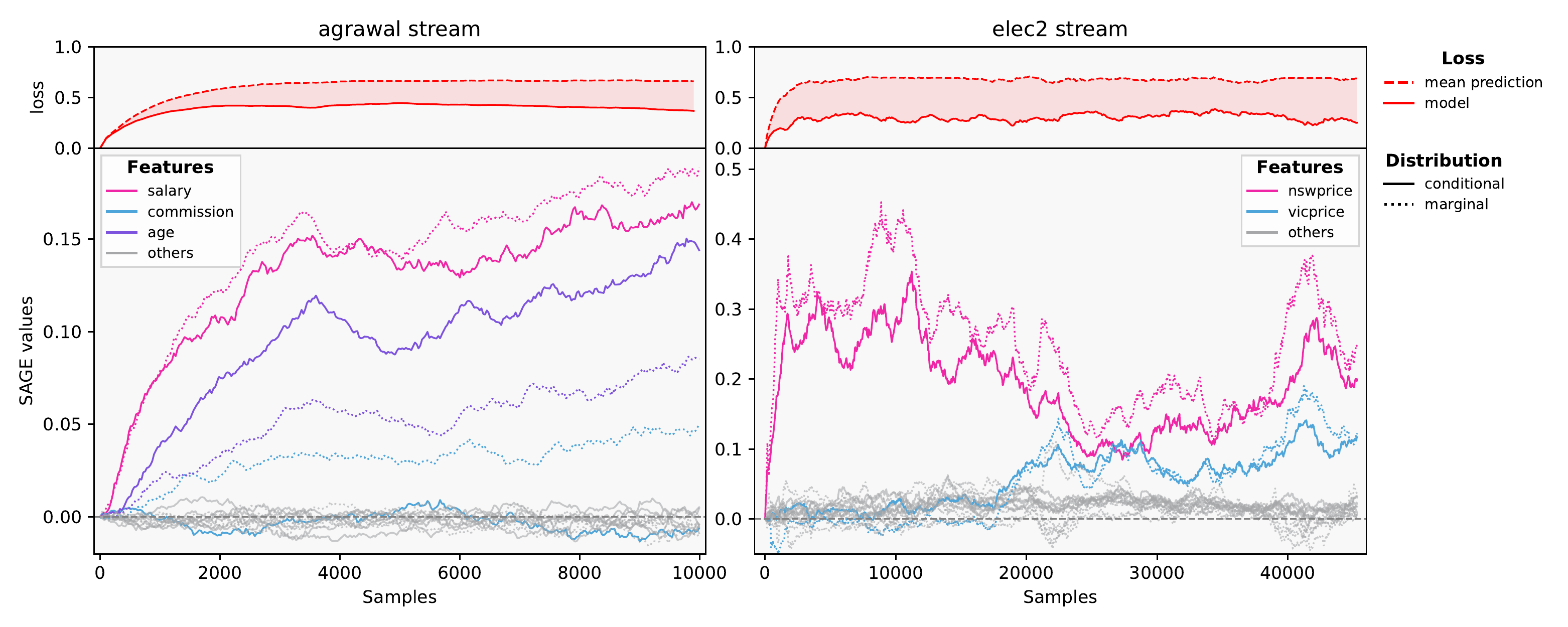}
    \caption{Comparison of iSAGE with conditional (dotted) and marginal (solid) for an \emph{agrawal} stream with known feature dependencies (left) and \emph{elec2} (right).}
    \label{fig:appendix_cond}
\end{figure}

Fig.~\ref{fig:appendix_cond} shows how the iSAGE values differ when we apply the observational feature removal mechanism compared to the interventional strategy.
We conduct this experiment on an \emph{agrawal} data stream with known feature dependencies and on the real-world data stream \emph{elec2}.
The classification function to be learned on the \emph{agrawal} stream is defined as 
\begin{align*}
    \text{class 1:\ } & ((X_{\text{age}} < 40) \land (50\,000 \leq X_{\text{salary}} \leq 100\,000 ))\ \lor \\ & ((40 \leq X_{\text{age}} < 60) \land (75\,000 \leq X_{\text{salary}} \leq 125\,000 ))\ \lor \\
    & ((X_{\text{age}} \geq 60) \land (25\,000 \leq X_{\text{salary}} \leq 75\,000 )).
\end{align*}

\end{document}